\documentclass[11pt]{article}

\usepackage[final]{acl}

\usepackage{times}
\usepackage{latexsym}

\usepackage[T1]{fontenc}

\usepackage[utf8]{inputenc}

\usepackage{microtype}

\usepackage{inconsolata}

\usepackage{graphicx}
\usepackage{xspace}
\usepackage[dvipsnames]{xcolor}
\usepackage{booktabs}
\usepackage{multirow}
\usepackage{tabularx}
\usepackage[table]{xcolor}
\usepackage{amsmath}
\usepackage{makecell}
\usepackage[framemethod=TikZ]{mdframed}
\usepackage{tcolorbox}
\tcbuselibrary{skins,breakable}
\usepackage{xcolor}
\usepackage{enumitem}

\newcommand{\cmark}{\textcolor{green!60!black}{\ding{51}}}
\newcommand{\xmark}{\textcolor{red!70!black}{\ding{55}}}

\definecolor{promptgray}{RGB}{170,170,170}
\definecolor{promptborder}{RGB}{190,190,190}

\newtcolorbox{promptbox}[1]{
    enhanced,
    colback=white,
    colframe=promptborder,
    colbacktitle=promptgray,
    coltitle=white,
    title={#1},
    fonttitle=\bfseries\Large,
    boxrule=0.8pt,
    arc=6pt,
    outer arc=6pt,
    left=12pt,
    right=12pt,
    top=12pt,
    bottom=12pt,
    titlerule=0pt,
    toptitle=8pt,
    bottomtitle=8pt,
    lefttitle=12pt,
    righttitle=12pt
}

\usepackage{pifont} 

\newcommand{\dataset}{VidOmni-Bench\xspace}
\title{VidOmni-Bench: A Benchmark for Fine-Grained Video Understanding via Spatio-Temporal Event Verification across Complexity and Duration}

\author{
    \textbf{Changbeen Kim}$^{1}$ \quad
    \textbf{Junwon Chang}$^{1}$ \quad
    \textbf{Kipyo Kim}$^{2}$ \quad
    \textbf{Risa Shinoda}$^{3}$ \\
    \textbf{Kuniaki Saito}$^{4}$ \qquad
    \textbf{Donghyun Kim}$^{1}$\thanks{Corresponding author.} \\
    \\
    \begin{tabular}{@{}c@{\hspace{1.5em}}c@{}}
        $^{1}$Korea University & $^{2}$Sungkyunkwan University \\
        $^{3}$The University of Tokyo & $^{4}$OMRON SINIC X Corporation
    \end{tabular} \\[0.4em]
    \raisebox{-0.15em}{\includegraphics[height=1.1em]{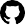}}\enspace
    \url{https://github.com/changbeendata/VidOmni-Bench}
}

\begin{document}
\maketitle
\begin{abstract}
While Video Large Language Models (Video-LLMs) have recently demonstrated strong performance, reliably evaluating their fine-grained video understanding remains challenging. Existing benchmarks often rely on question answering or ground-truth caption matching, where models may succeed through superficial cues and incomplete annotations. To this end, we introduce VidOmni-Bench, a benchmark that requires models to verify whether each event in dense video captions is supported by the video. VidOmni-Bench consists of 500 videos spanning five complexity types and diverse durations from 4 seconds to 90 minutes. After collecting videos along these axes, we use diverse Video-LLMs to generate dense captions and obtain human-verified sentence-level labels, where sentences containing incorrect events serve as hard negatives for evaluation. Our experiments on VidOmni-Bench reveal three key findings: (i) Video-LLMs frequently generate hallucinated descriptions in dense video captioning; (ii) they also struggle as verifiers, failing to reliably detect plausible but incorrect event descriptions; and (iii) model weaknesses vary across video complexity and duration, revealing diverse, model-specific bottlenecks in current Video-LLMs.

\end{abstract}

\section{Introduction}

Video Large Language Models (Video-LLMs)~\cite{tang2025videoa,bai2025qwen3,wang2025internvl3,yang2025kwai} have demonstrated strong capabilities in video understanding tasks such as question answering and video retrieval. However, reliably evaluating their fine-grained video understanding remains challenging~\cite{kong2025tuna,cai2024temporalbench}, as videos vary widely in duration and visual dynamics. A comprehensive evaluation should therefore assess whether models can interpret diverse video evidence across different temporal scales and visual aspects, including object states, actions, background changes, and scene transitions.

\begin{figure} 
    \centering
    \includegraphics[width=1.0\linewidth]{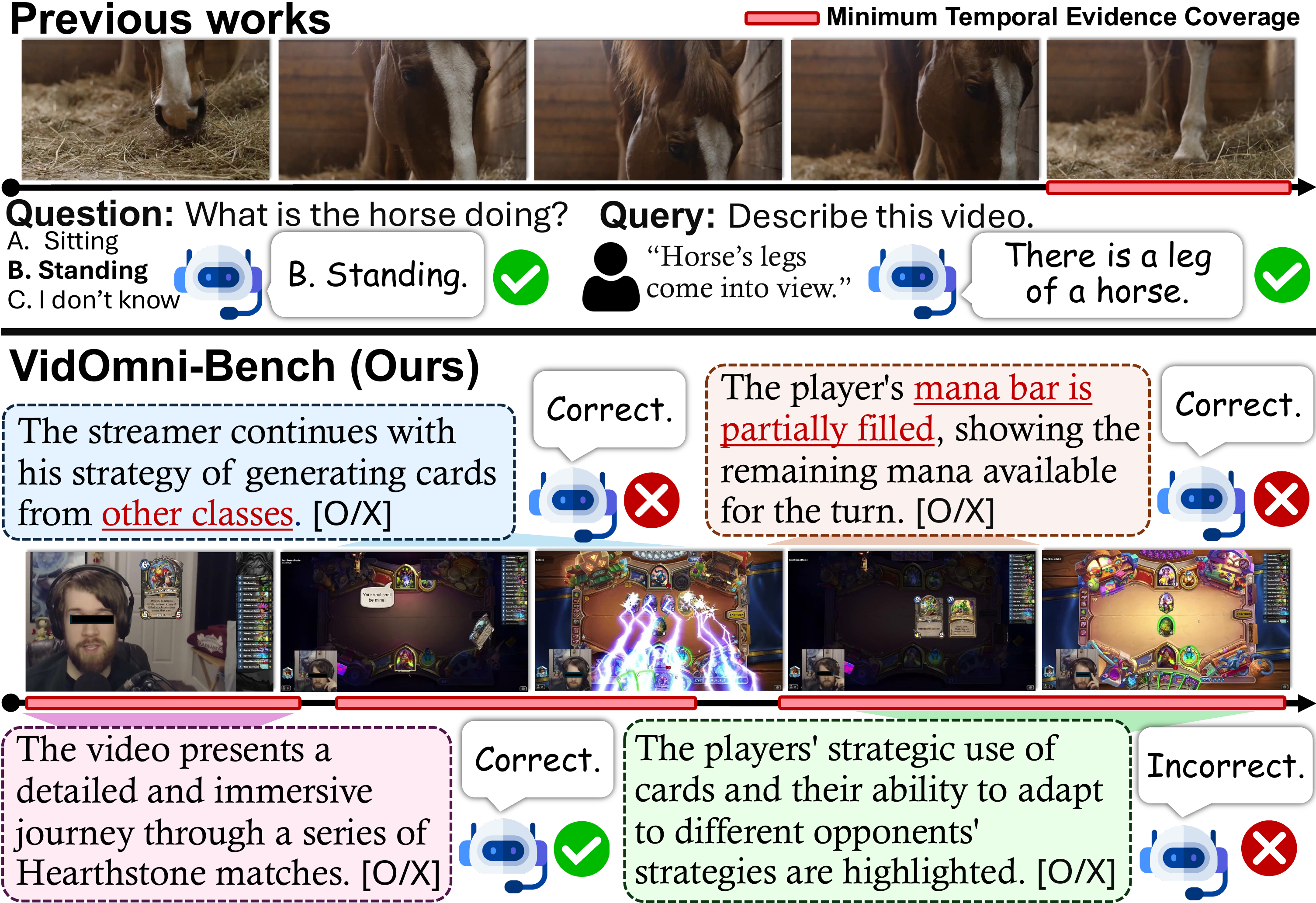}
    \vspace{-2mm}
    \caption{\textbf{Top:} Previous benchmarks typically rely on MCQA or sparse GT captions, allowing models to succeed with limited temporal evidence. \textbf{Bottom:} In \dataset, models must verify each event in dense captions against the video, enabling a more rigorous assessment of fine-grained video understanding.}
    \vspace{-2mm}
    \label{fig:figure_1}
\end{figure}

Recent temporal and long-form video benchmarks mostly rely on Multiple-Choice Question Answering (MCQA)~\cite{Fu_2025_CVPR,fang2024mmbench,tan2025allvb} or Ground-Truth (GT) caption matching~\cite{rawal2025argus,liu2025capability,lee2025noah}. MCQA-based evaluations may allow models to rely on localized evidence, or textual shortcuts without temporal reasoning~\cite{feng2025breaking}. Caption-matching-based benchmarks can assess video understanding more rigorously, but their focus on relatively short videos and human-written GT captions limits scalable dense event-level coverage across diverse video durations. Consequently, existing benchmarks do not directly evaluate whether Video-LLMs can understand fine-grained video events by integrating broad temporal evidence across diverse durations.

\begin{figure*} 
    \centering
    \includegraphics[width=1.0\linewidth]{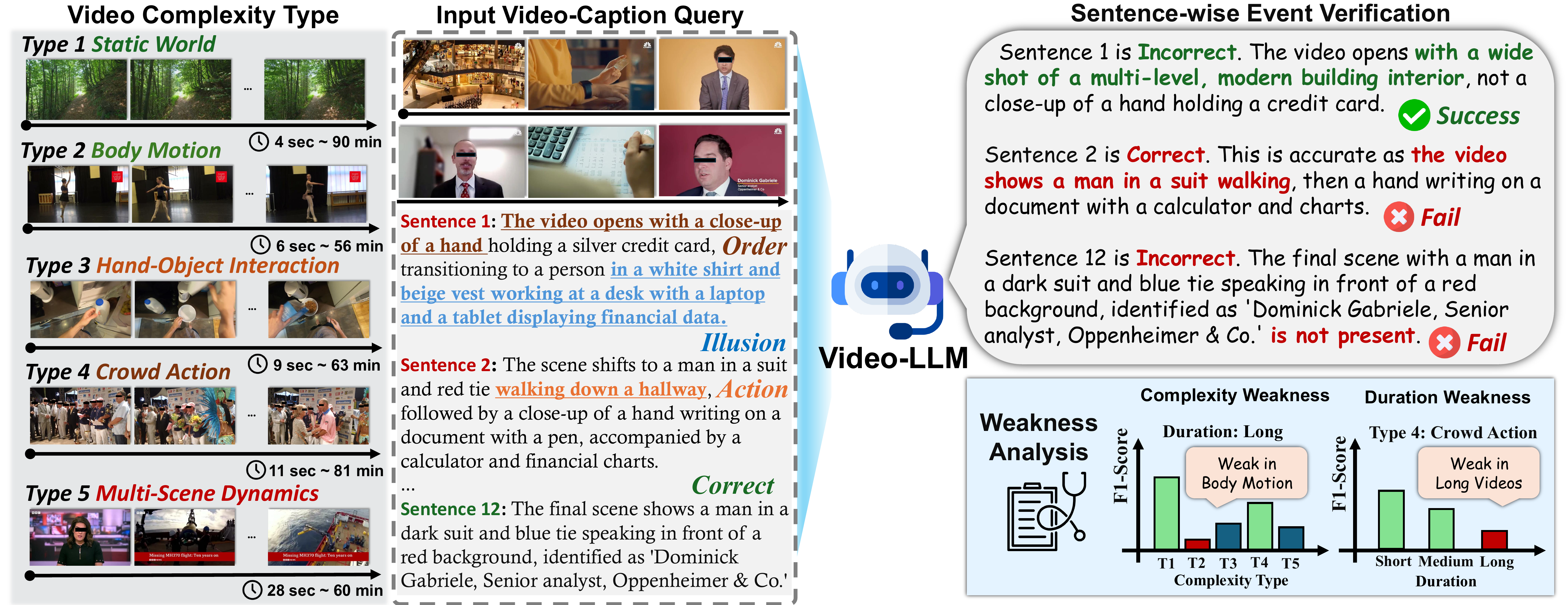}
    \vspace{-2mm}
    \caption{We introduce VidOmni-Bench, a novel benchmark designed to evaluate how well Video-LLMs distinguish between correct and incorrect events across five types of spatio-temporal complexity and diverse video durations. This structured framework enables a rigorous assessment of whether Video-LLMs can truly ground their reasoning in the visual evidence across diverse scales.}
    \vspace{-2mm}
    \label{fig:main}
\end{figure*}

To address these gaps, we introduce \dataset, a sentence-wise event verification benchmark where a model verifies whether each sentence in dense video captions is supported by the video. As shown in Figure~\ref{fig:main}, \dataset consists of 500 videos curated across five complexity types and diverse durations from 4 seconds to 90 minutes. For each video, we use a diverse set of Video-LLMs to generate dense captions, and human annotators verify each sentence to produce sentence-level labels. Hallucinated or incorrect sentences are then used as hard negatives for evaluation, resulting in a benchmark designed to assess event verification across diverse types of video complexity and duration. As shown in Table~\ref{tab:benchmark_comparison}, \dataset requires models to inspect substantially broader portions of the video than existing benchmarks, indicating that VidOmni-Bench cannot be solved through superficial cues. Furthermore, Figure~\ref{fig:vidomni_validity} shows that the captions in \dataset provide dense and complex event descriptions across the five complexity types. These results support \dataset as a challenging benchmark for fine-grained video understanding.

Our experiments on \dataset reveal several critical limitations of current Video-LLMs. First, Video-LLMs frequently produce hallucinated descriptions during dense video captioning, indicating unreliable video-grounded generation. Second, they also struggle as verifiers, failing to reliably detect plausible but incorrect event descriptions within these captions. Third, model weaknesses vary substantially across video complexity and duration, revealing diverse, model-specific bottlenecks rather than a single uniform failure mode. These findings indicate that current Video-LLMs still lack robust, temporally grounded event understanding needed to faithfully interpret dense and diverse video content.

\begin{figure*}
    \centering
    \includegraphics[width=0.9\linewidth]{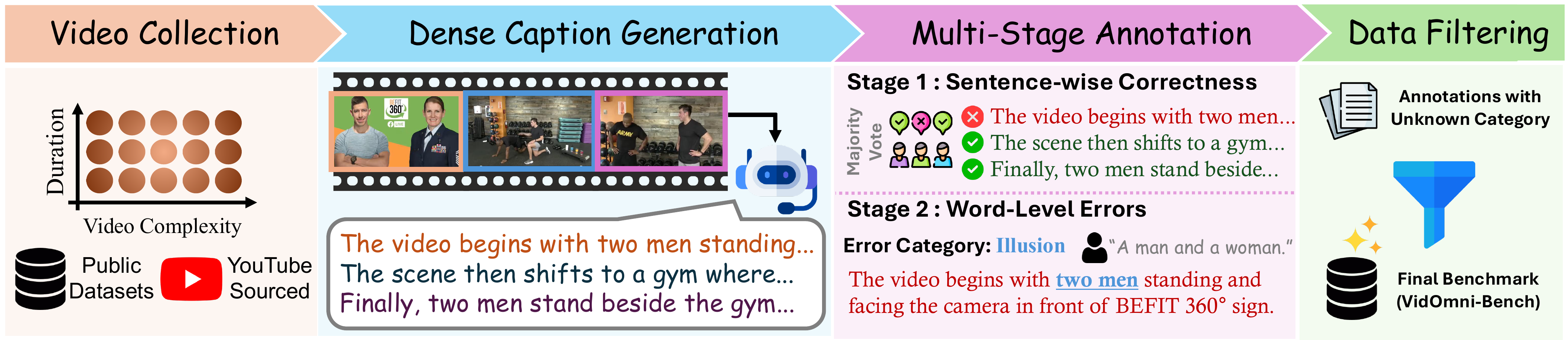}
    \vspace{-2mm}
    \caption{\textbf{Construction pipeline of VidOmni-Bench.} We collect videos from public datasets and YouTube along with two axes. Then, we utilize diverse Video-LLMs for dense caption generation. These video and dense caption pairs then undergo a multi-stage annotation phase for sentence-wise error (hallucination) labeling, where annotators label sentence-level errors to construct hard negatives. Finally, the data filtering stage for a robust evaluation.}
    \vspace{-2mm}
    \label{fig:anno_pipe}
\end{figure*}

\section{Related Works}

\noindent\textbf{Video Large Language Models.}
Following the success of Large Language Models (LLMs)~\cite{yang2025qwen3,grattafiori2024llama}, multimodal understanding has expanded from static images~\cite{liu2023visual,chen2024internvl} to dynamic videos~\cite{li2024llava,li2024llama,chen2024sharegpt4video,bai2025qwen3}. Recent Video-LLMs can generate detailed video descriptions, but their outputs often contain subtle hallucinations~\cite{gao2025exploring}. We use these generated descriptions for sentence-wise event verification, with human annotators labeling hallucinated sentences as hard negatives.

\noindent\textbf{Video Understanding Benchmarks.}
Recent benchmarks have advanced the evaluation of temporal and long-form video understanding~\cite{ben2025herbench,ha2026narrativetrack,li2026timeblind,cai2024temporalbench,kong2025tuna}, but many remain based on MCQA formats. Such positive-selection formats can be affected by language priors or partial visual evidence~\cite{feng2025breaking}, and provide limited evidence of whether models can reject plausible but incorrect events. \dataset instead evaluates dense event verification beyond sparse question-answer pairs.

\noindent\textbf{Hallucination in Video Understanding.}
Several studies evaluate video captioning and hallucination detection by comparing model outputs with human-written GT captions~\cite{liu2025capability,rawal2025argus,lee2025noah}. While useful for assessing factual consistency, such sparse references limit dense event-level coverage across diverse domains and durations. Motivated by~\citet{saito2026haldec}, \dataset uses Video-LLM-generated dense descriptions and human-verified negatives to analyze hallucination patterns across video complexity and duration.

\begin{figure}
    \centering
    \includegraphics[width=0.85\linewidth]{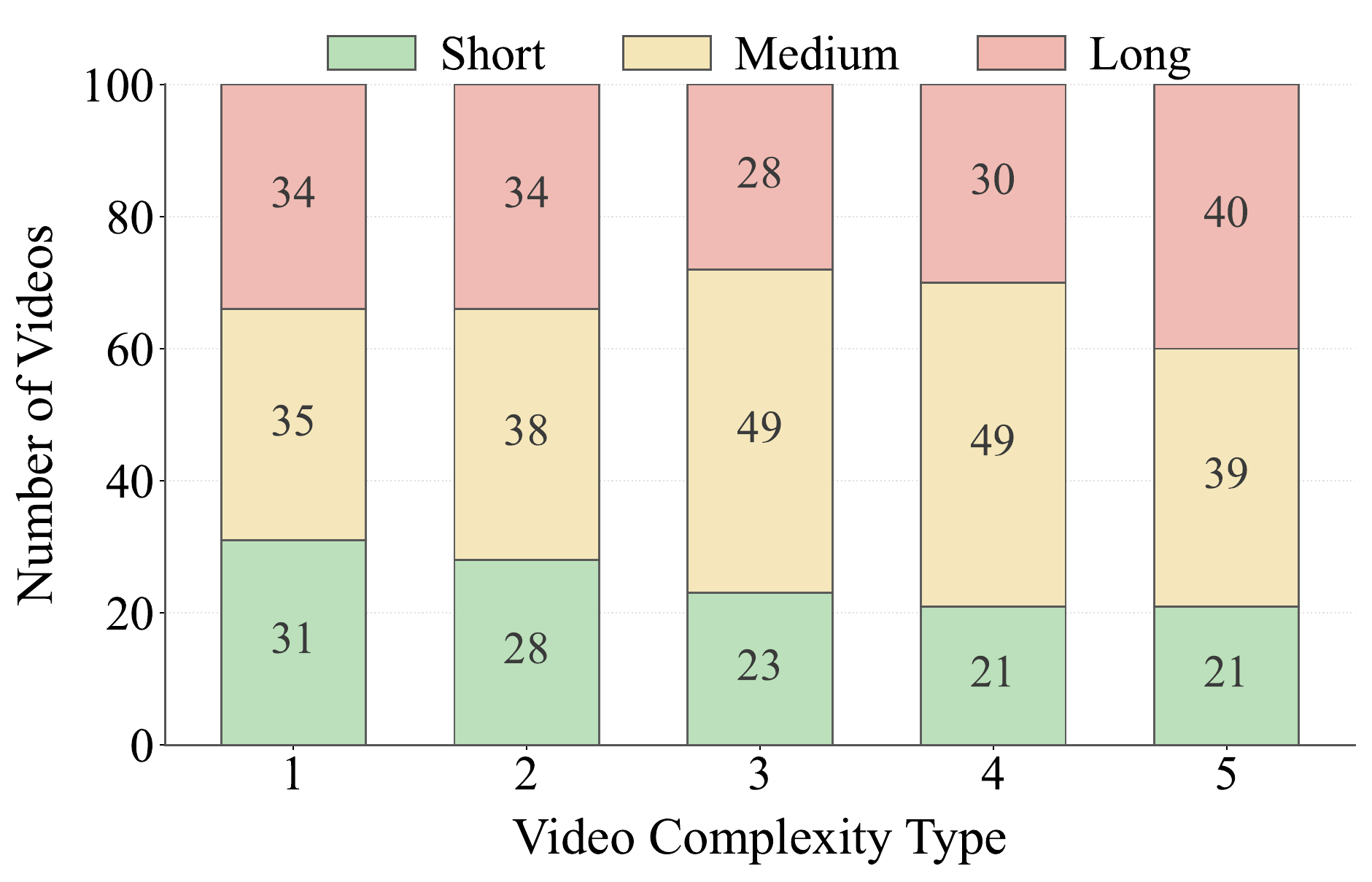}
    \vspace{-2mm}
    \caption{Our dataset consists of 500 videos, with exactly 100 videos allocated to each of the five complexity types. Within each type, we maintain a balanced distribution across Short, Medium, and Long duration.}
    \vspace{-2mm}
    \label{fig:duration}
\end{figure}

\section{VidOmni-Bench}
To evaluate the fine-grained understanding of Video-LLMs, we introduce a sentence-wise event verification task. This task requires models to act as rigorous verifiers, determining both the factual occurrence of described events and their temporal consistency within the video. To provide a granular diagnosis of model capabilities, we constructed VidOmni-Bench by carefully disentangling spatio-temporal complexity from video duration. Rather than relying on human-written GT captions, we generate dense captions using diverse Video-LLMs and verify them at the sentence level. The following sections define the task, describe the video collection and annotation process, and present statistical analysis. Overall construction process of \dataset is illustrated in Figure~\ref{fig:anno_pipe}.

\subsection{Task Definition}
Sentence-wise event verification evaluates whether each sentence in a video caption is factually supported by the video. Formally, given a video $V$ and an $n$-sentence caption $C = \{s_1, s_2, \dots, s_n\}$, the goal is to predict binary labels $y = \{y_1, y_2, \dots, y_n\}$. For each sentence $s_i$, we define $y_i \in \{0,1\}$ as:
\begin{equation}
y_i = 
\begin{cases} 
0, & \text{if } s_i \text{ is factually supported by } V, \\
1, & \text{otherwise}.
\end{cases}
\end{equation}
Thus, a Video-LLM must verify each sentence against the video evidence and determine whether it contains hallucinated or incorrect content.

\subsection{Video Collection}
To enable a granular diagnosis of Video-LLMs, we curate videos by controlling spatio-temporal complexity and video duration as two separate axes. We collect videos from diverse public datasets, covering egocentric actions~\cite{grauman2022ego4d,damen2018scaling,sener2022assembly101}, anomaly events~\cite{Sultani_2018_CVPR,Wu2020not}, and long-form content~\cite{caba2015activitynet,Fu_2025_CVPR}. We further supplement these sources with targeted YouTube crawling to fill underrepresented combinations of duration and complexity. For crawled videos, we only retain content released under Creative Commons licenses.

Our collection is stratified into five predefined spatio-temporal complexity types: (1) Static World, (2) Body Motion, (3) Hand-Object Interaction, (4) Crowd Action, and (5) Multi-Scene Dynamics. As shown in Figure~\ref{fig:duration}, each complexity type is balanced across three duration brackets: short ($\leq$ 3 min), medium (4--30 min), and long ($\geq$ 30 min). This design enables us to analyze the effects of complexity and duration separately. The resulting VidOmni-Bench contains 500 videos, 433 of which include audio tracks. Detailed definitions of complexity types and data sources are provided in the Appendix~\ref{appendix_dataset}.

\begin{table*}[t]
\centering
\small
\renewcommand{\arraystretch}{1.2}
\resizebox{0.90\textwidth}{!}{\begin{tabular*}{\textwidth}{@{\extracolsep{\fill}} l ccccc @{}}
\toprule
\textbf{} & \textbf{KeyeVL 1.5} & \textbf{InternVL 3.5} & \textbf{Qwen3-VL} & \textbf{video-SALMONN 2+} & \textbf{Gemini 3} \\ \midrule
Params. & 8B & 38B & 32B & 72B & N/A \\
Total Sentences & 18974 & 5542 & 11868 & 18974 & 9485 \\
Video Incorrect Rate (\%) & 92.6 & \textbf{85.6} & 94.0 & 92.2 & 91.1 \\
Incorrect / Caption (\%) & 40.6 & 39.5 & 26.5 & 44.9 & \textbf{24.1} \\ \bottomrule
\end{tabular*}}
\caption{\textbf{Statistics of incorrect sentence in generated captions.} \dataset provides a large set of annotated incorrect sentences for evaluating sentence-wise event verification. Sentences labeled as Unknown are excluded.}
\label{tab:captioning_stats}
\end{table*}

\subsection{Annotation Pipeline}
To overcome the limited coverage of human-written GT captions, we use modern Video-LLMs to generate dense descriptions and repurpose their human-verified hallucinations as hard negatives. We use five Video-LLMs with diverse architectures and parameter scales: InternVL3.5~\cite{wang2025internvl3} (38B), Qwen3-VL~\cite{bai2025qwen3} (32B), video-SALMONN 2+~\cite{tang2025videob} (72B), Keye-VL 1.5~\cite{yang2025kwai} (8B), and Gemini 3 Flash~\cite{google2025gemini}. For each video, one of five specialized chat templates (e.g., ``Describe the full content of the video in detail.'') is randomly assigned to each model, producing 2,500 video-caption pairs in total. Using multiple models and prompts diversifies the generated descriptions and reduces dependence on the output of a single model.

These raw captions serve as the basis for a two-stage human annotation process. In the first stage, five independent annotators verify the factual correctness of each sentence and label it as Correct, Incorrect, or Unknown. Final labels are determined by majority vote, and Unknown cases are excluded to maintain reliable evaluation.

In the second stage, we perform word-level error annotation for sentences labeled as Incorrect. Annotators identify the incorrect words within each sentence and assign each error to one of ten categories: five related to static-frame understanding (Object, Relation, Location, Direction, and Illusion) and five related to temporal understanding (Action, Order, Causality, Frequency, and State Change). The resulting human-verified incorrect sentences serve as hard negatives for sentence-wise event verification. Details for dense captioning, and annotation process are provided in the Appendix~\ref{appendix_dataset}.
\begin{figure}
    \centering
    \includegraphics[width=1.0\linewidth]{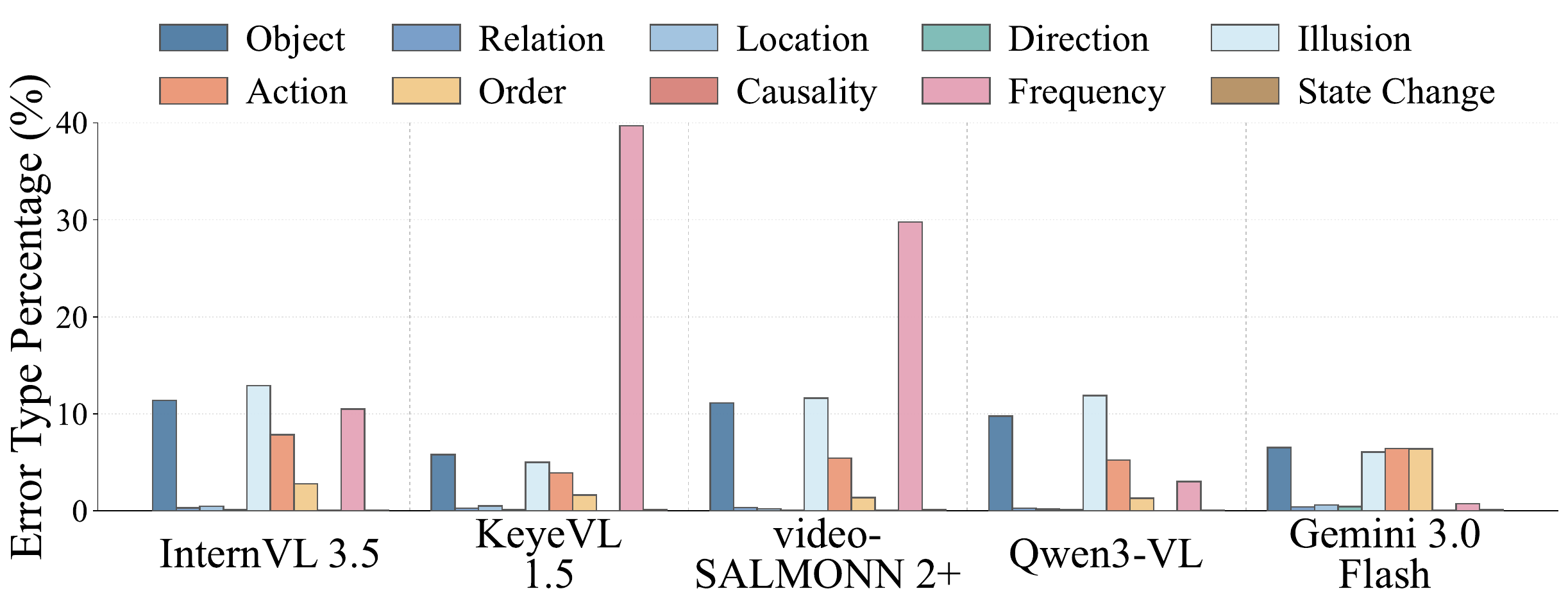}
    \caption{\textbf{Word-level error statistics per category.} The consistent high frequency of temporal errors in dense video captioning highlights a fundamental bottleneck in video understanding.}
    \label{fig:error_stats}
\end{figure}

\begin{figure}
    \centering
    \includegraphics[width=1.0\linewidth]{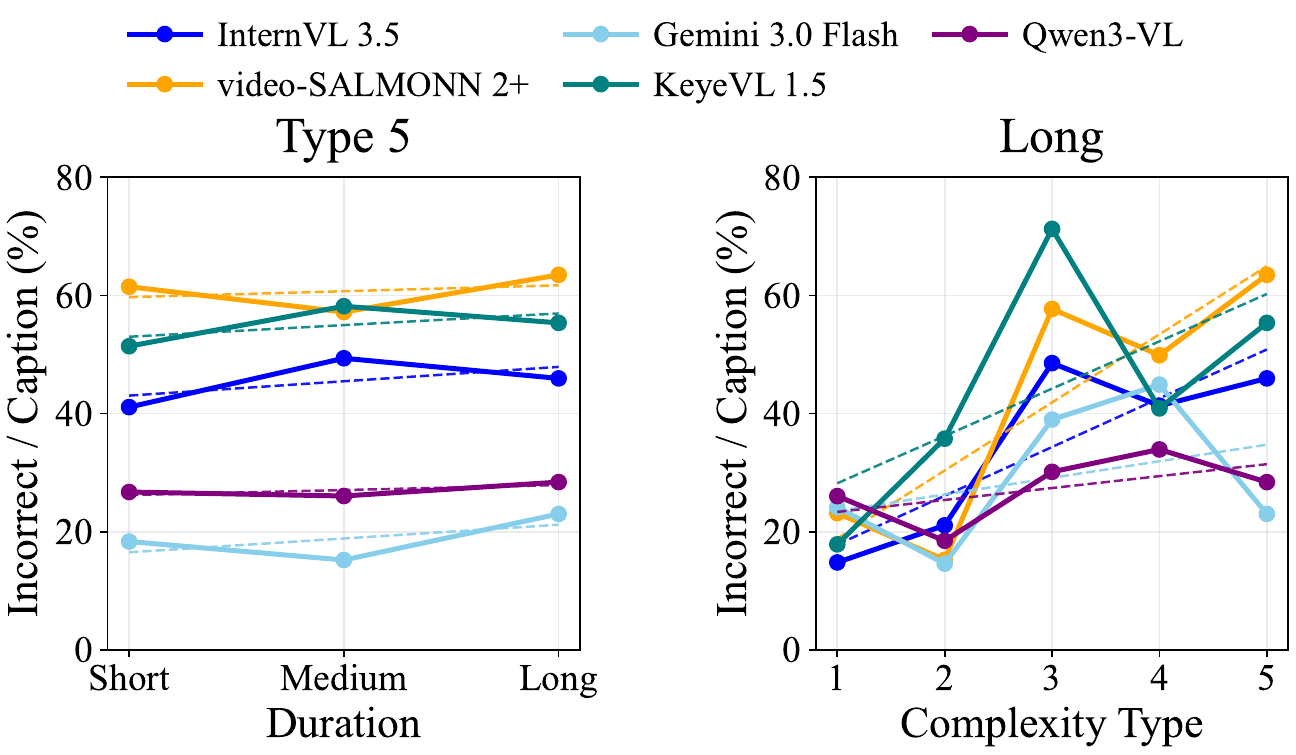}
    \caption{\textbf{Effect of video complexity and duration on hallucination.} Plots show the average sentence-level Incorrect Rate per caption. \textbf{Left}: across durations for Type 5 videos. \textbf{Right}: across complexity types within Long videos. Dashed lines indicate overall trends.}
    \label{fig:challenging_in_captioning}
\end{figure}

\subsection{Data Statistics and Analysis}
\label{subsec:annotation_pipeline}

\noindent\textbf{Basic statistics.} 
Table~\ref{tab:captioning_stats} summarizes VidOmni-Bench, which contains 64K annotated sentences. We report the Video Incorrect Rate, defined as the proportion of video-caption pairs in which the generated caption contains at least one hallucinated sentence, along with the average sentence-level Incorrect Rate per caption. The results show that Video-LLMs frequently generate hallucinated captions, with Incorrect rates remaining high across all models, thereby providing a sufficient pool of hard negatives for sentence-wise event verification.

\noindent\textbf{Error category distribution.}
As shown in Figure~\ref{fig:error_stats}, temporal errors such as Action, Order, and Frequency account for a substantial portion of hallucinations, suggesting that current Video-LLMs struggle with fine-grained temporal reasoning. Figure~\ref{fig:main} further shows that a sentence can appear plausible in isolation but become Incorrect when verified against the broader video context.

\noindent\textbf{Effect of complexity and duration on dense captioning.}
By controlling video complexity and duration, we analyze hallucination patterns in dense captioning (Figure~\ref{fig:challenging_in_captioning}). The sentence-level Incorrect Rate per caption generally increases with complexity type within each duration bracket, while Type 5 remains consistently high across all durations. This suggests that hallucinations are driven not only by video length, but also by the difficulty of reasoning over complex event dynamics.

\begin{figure}
    \centering
    \includegraphics[width=\linewidth]{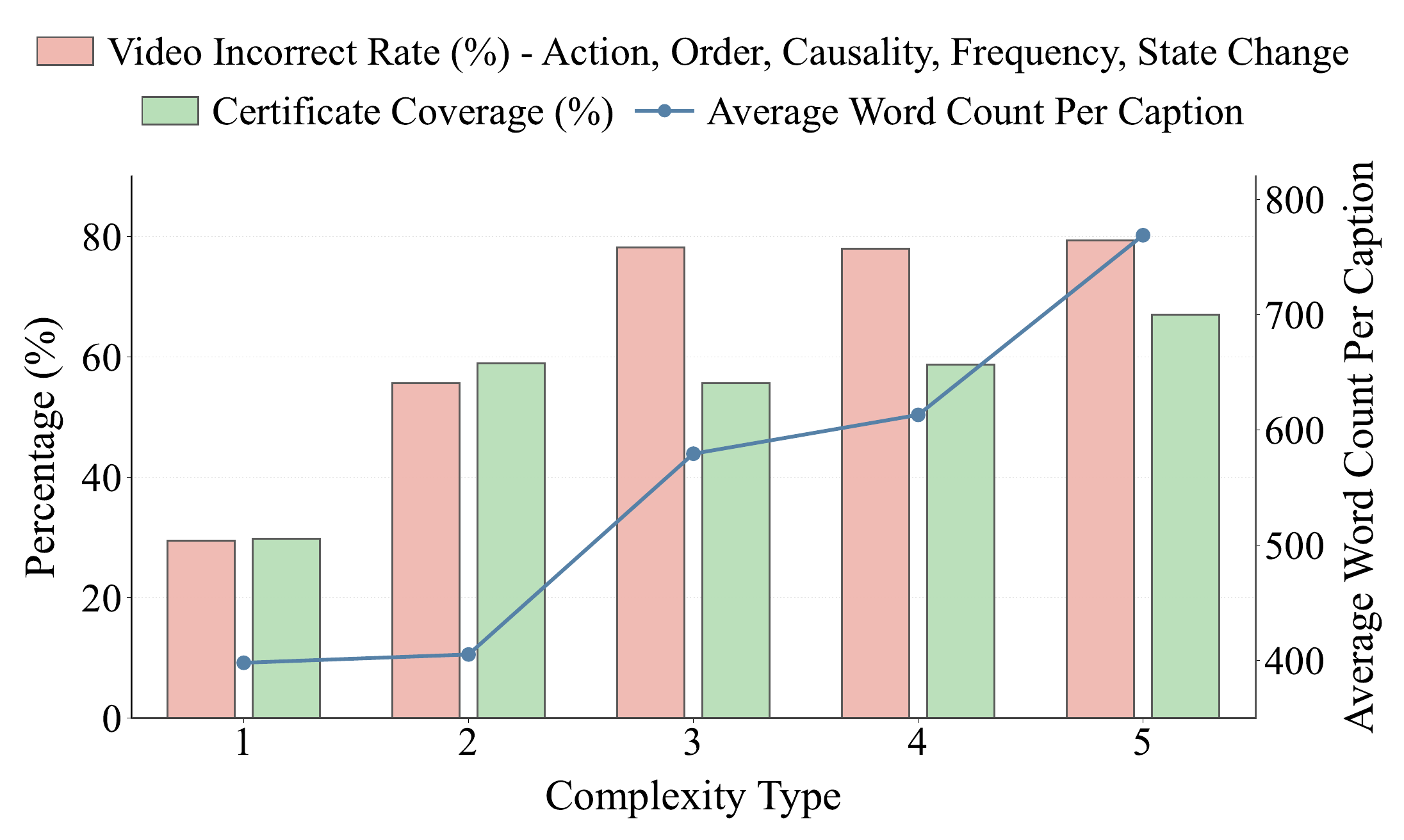}
    \caption{\textbf{Caption characteristics across complexity types.} Higher-complexity videos lead to longer captions and a higher Video Incorrect Rate for temporal error categories, while also requiring broader video evidence in sentence-wise event verification of \dataset.}
    \label{fig:vidomni_validity}
\end{figure}

\noindent\textbf{Caption characteristics differ across complexity types.}
We analyze how caption properties and verification requirements vary across the five complexity types in \dataset. Figure~\ref{fig:vidomni_validity} shows that higher complexity is associated with longer captions and higher Video Incorrect Rates over temporal error categories(Action, Order, Causality, Frequency, and State Change). We further report Certificate Coverage following Video-MME~\cite{Fu_2025_CVPR}, showing that sentence-wise event verification generally requires broader video evidence as complexity increases. Details for calculating Certificate Coverage are provided in the 
Appendix~\ref{additional_analysis}.

\noindent\textbf{Comparison with existing benchmarks.}
Figure~\ref{fig:figure_1} and Table~\ref{tab:benchmark_comparison} position VidOmni-Bench among existing video understanding benchmarks. Moving beyond MCQA-based evaluation~\cite{Fu_2025_CVPR, ben2025herbench,li2026timeblind}, VidOmni-Bench performs sentence-wise event verification over dense descriptions. Compared with benchmarks based on human-written GT caption matching~\cite{rawal2025argus,lee2025noah, liu2025capability, ma2025videoeval}, \dataset spans broader video durations. Its higher Certificate Coverage further shows that verification requires broader video evidence, while the Appendix~\ref{additional_analysis} shows that \dataset captions are more complex than those in prior benchmarks.

\begin{table}[t]
\centering
\renewcommand{\arraystretch}{1.1}
\resizebox{\columnwidth}{!}{%
\begin{tabular}{l c c c}
\toprule
\textbf{Benchmark} & \textbf{Evaluation Task} & \textbf{Duration} & \textbf{\makecell{Cert. \\ Cov. (\%)}} \\ \midrule
TimeBlind \cite{li2026timeblind} & MCQA & $\leq$ 25s & 32.3 \\
NOAH~\cite{lee2025noah} & GT Matching / MCQA & $\leq$ 12m & 30.8 \\
ARGUS~\cite{rawal2025argus} & GT Matching & $\leq$ 3m & 25.0 \\
HERBench~\cite{ben2025herbench} & MCQA & $\leq$ 6m & 24.8 \\
Video-MME~\cite{Fu_2025_CVPR} & MCQA & $\leq$ 60m & 3.4 \\
CAPAbility~\cite{liu2025capability} & GT Matching / MCQA & $\leq$ 50s & 2.0 \\ 
VideoEval-Pro~\cite{ma2025videoeval} & GT Matching & $\geq$ 30m & 0.9 \\ \midrule
\textbf{VidOmni-Bench (Ours)} & \textbf{Sentence Verification} & \textbf{$\leq$ 90m} & \textbf{54.0} \\ \bottomrule
\end{tabular}
}
\caption{\textbf{Comparison with other benchmarks in task design and temporal evidence.}
\dataset evaluates sentence-wise event verification over longer videos and requires broader temporal evidence, indicated by higher Certificate Coverage (Cert. Cov.).}
\label{tab:benchmark_comparison}
\end{table}

\section{Experiments}
In this section, we evaluate a wide range of Video-LLMs on \dataset to assess their fine-grained video understanding. We first present quantitative results, followed by a detailed analysis of model performance across different dimensions. In summary, our analysis yields five key findings:
(i) \dataset reveals verification weaknesses not captured by existing benchmarks, showing that models often fail to consistently identify hallucinated sentences;
(ii) scaling model size mainly improves verification of static-frame-related errors, whereas temporal ordering and causal reasoning remain major bottlenecks;
(iii) current Video-LLMs exhibit a distinct self-preference bias, showing unreliable verification behavior when judging their own generated sentences;
(iv) audio often provides verification-relevant information complementary to visual evidence, and models that effectively incorporate these cues achieve stronger performance; and
(v) model vulnerabilities are multidimensional, with performance depending on model-specific sensitivity to both complexity and duration.

\begin{table*}[t]
\centering
\small
\setlength{\tabcolsep}{2pt}
\renewcommand{\arraystretch}{1.3}
\resizebox{0.9\textwidth}{!}{
\begin{tabularx}{\textwidth}{l @{\extracolsep{\fill}} c c c c c c c}
\toprule
\multirow{2}{*}{\textbf{Video-LLM}} & \multirow{2}{*}{\textbf{Modality}} & \textbf{Video-MME} & \textbf{Video-MMMU} & \textbf{LVBench} & \multicolumn{3}{c}{\textbf{VidOmni-Bench}} \\ \cmidrule(lr){3-3} \cmidrule(lr){4-4} \cmidrule(lr){5-5} \cmidrule(lr){6-8}
 & & QA-Acc. (\%) & QA-Acc. (\%) & QA-Acc. (\%) & Precision (\%) & Recall (\%) & F1-Score (\%) \\ \midrule

\rowcolor{gray!15} \multicolumn{8}{@{} l @{}}{\quad \textit{Open-Source Models}} \\
Keye-VL 1.5 8B & $v$ & 73.0 & 66.0 & 35.1$^\dagger$ & 19.1 & 8.4 & 9.6 \\ \hline
InternVL 3.5 2B & $v$ & 58.4 & 41.8$^\dagger$ & 30.5$^\dagger$ & 6.0 & 1.8 & 2.3 \\
InternVL 3.5 8B & $v$ & 66.0 & 48.9$^\dagger$ & 34.2$^\dagger$ & 26.4 & 13.9 & 14.7 \\
InternVL 3.5 14B & $v$ & 67.9 & 54.8$^\dagger$ & 35.1$^\dagger$ & 32.3 & 18.1 & 19.4 \\
InternVL 3.5 38B & $v$ & 70.9 & 57.9$^\dagger$ & 38.1$^\dagger$ & 25.0 & 12.7 & 14.1 \\ \hline
Qwen3-VL 2B & $v$ & 61.9 & 41.9 & 47.4 & 5.5 & 4.2 & 3.6 \\
Qwen3-VL 4B & $v$ & 69.3 & 56.2 & 56.2 & 22.9 & 11.7 & 12.7 \\
Qwen3-VL 8B & $v$ & 71.4 & 65.3 & 58.0 & 25.9 & 13.6 & 14.7 \\
Qwen3-VL 32B & $v$ & 76.6 & 71.9 & 63.8 & 35.4 & 19.6 & 21.7 \\ \hline
video-SALMONN 2+ 7B & $v$ & 67.8$^\dagger$ & 46.0$^\dagger$ & 49.7 & 1.3 & 1.1 & 1.1 \\
video-SALMONN 2+ 7B & $v, a$ & 73.4 & 50.2$^\dagger$ & 47.8$^\dagger$ & 0.1 & 0.2 & 0.1 \\
video-SALMONN 2+ 72B & $v$ & 74.0$^\dagger$ & 62.0$^\dagger$ & 55.5 & 8.3 & 5.4 & 5.8 \\
video-SALMONN 2+ 72B & $v, a$ & 79.7 & 60.8$^\dagger$ & 55.2$^\dagger$ & 5.7 & 3.4 & 3.7 \\ \midrule

\rowcolor{gray!15} \multicolumn{8}{@{} l @{}}{\quad \textit{Closed-Source Models}} \\
Gemini 3 Flash & $v$ & 73.2$^\dagger$ & 80.3$^\dagger$ & 74.3$^\dagger$ & 54.6 & 37.2 & 39.8 \\
Gemini 3 Flash & $v, a$ & 86.1$^\dagger$ & 75.3$^\dagger$ & 76.9$^\dagger$ & \textbf{54.7} & \textbf{39.9} & \textbf{41.7} \\
GPT 5 mini & $v$ & 71.0 & 56.7 & 57.8$^\dagger$ & 39.3 & 33.2 & 28.9 \\ \bottomrule
\end{tabularx}
}
\caption{\textbf{Main results on VidOmni-Bench.} We report verification Precision, Recall, and F1-Score for \dataset, alongside QA Accuracy (QA-Acc) for Video-MME, Video-MMMU, and LVBench. Models are evaluated under video-only ($v$) or video+audio ($v,a$) settings. $\dagger$ denotes reproduced performance.}
\label{tab:main_results}
\end{table*}

\begin{figure}
    \centering
    \includegraphics[width=1.0\linewidth]{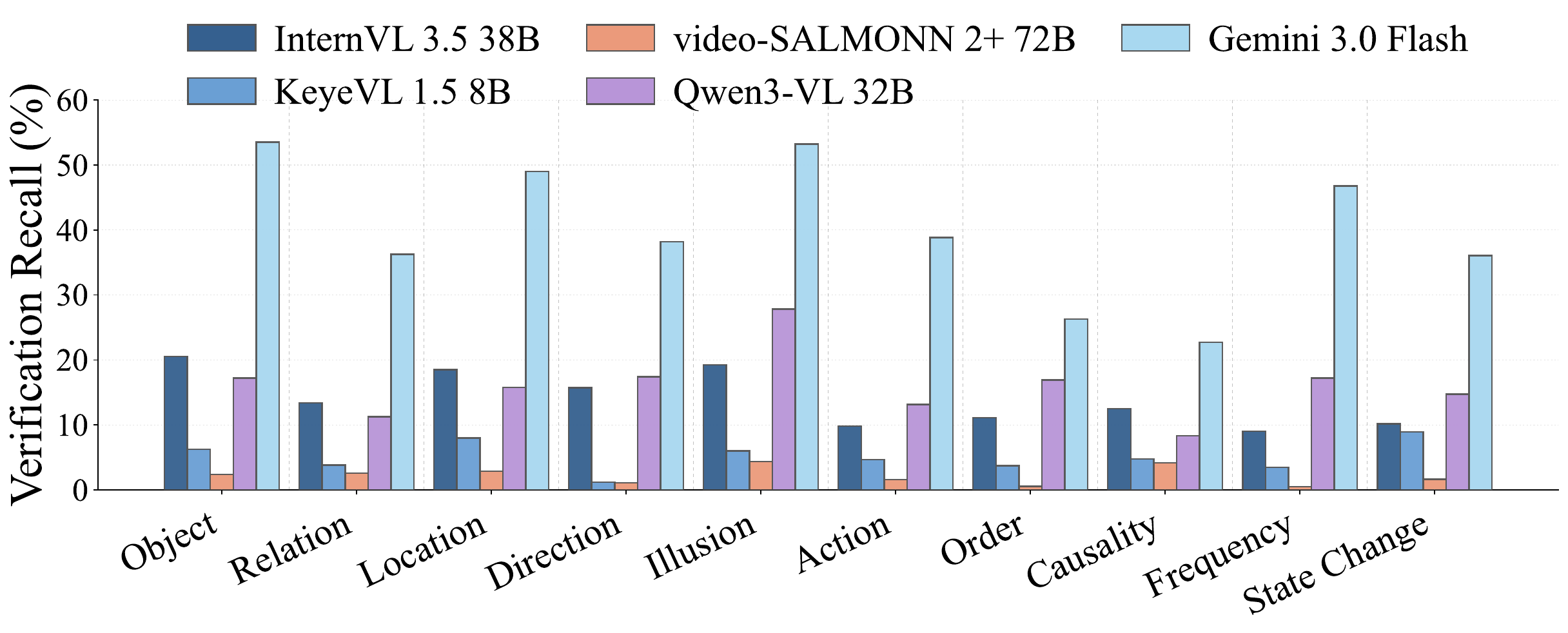}
    \caption{\textbf{Performance of Video-LLMs across error categories.} This visualization highlights the relative strengths and weaknesses of current Video-LLMs in detecting specific static and temporal discrepancies.}
    \label{fig:label_recall}
\end{figure}

\begin{figure*}
    \centering
    \includegraphics[width=1.0\linewidth]{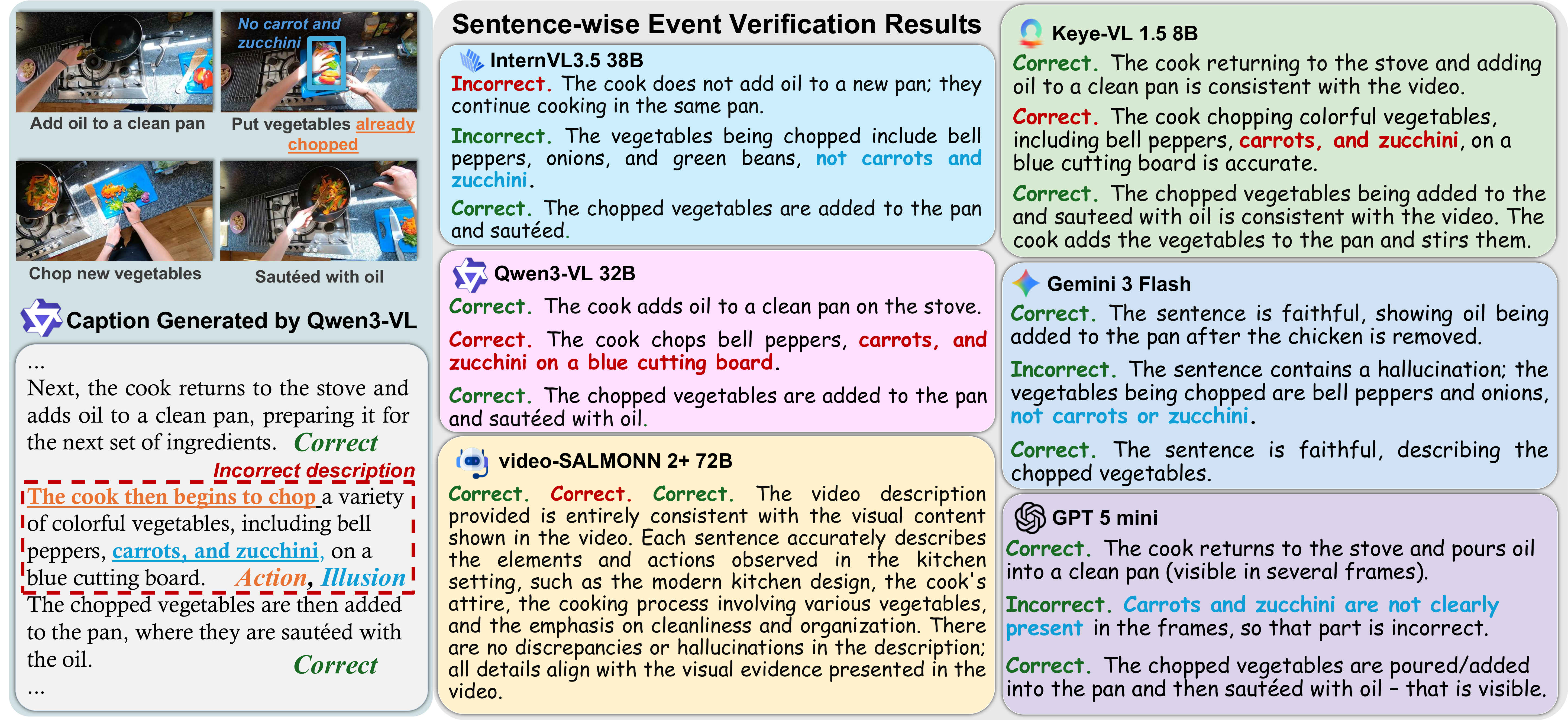}
    \caption{\textbf{Qualitative examples of sentence-wise event verification on a Qwen3-VL caption.} Video-LLMs may produce the same prediction while relying on different reasoning, revealing varying degrees of video understanding.}
    \label{fig:example_haldet}
\end{figure*}

\subsection{Experimental Setup}

\noindent\textbf{Benchmark coverage and evaluation protocols.}
We evaluate Video-LLMs on \dataset together with three existing video understanding benchmarks: Video-MME~\cite{Fu_2025_CVPR}, Video-MMMU~\cite{hu2025video}, and LVBench~\cite{wang2025lvbench}. Since these benchmarks are formulated as Question Answering (QA) tasks, we report QA accuracy following their original evaluation protocols. In contrast, \dataset is formulated as a sentence-wise event verification task, where sentences labeled Incorrect are treated as the detection target. For the $i$-th video-caption pair ($i = 1, \dots, N$), we compute Precision, Recall, and F1-Score over all sentences within the pair:
\begin{gather}
\text{Precision}_i = \frac{\text{TP}_i}{\text{TP}_i+\text{FP}_i}, \quad
\text{Recall}_i = \frac{\text{TP}_i}{\text{TP}_i+\text{FN}_i}, \nonumber \\[6pt]
\text{F1-Score}_i = \frac{2 \cdot \text{Precision}_i \cdot \text{Recall}_i}{\text{Precision}_i+\text{Recall}_i}.
\end{gather}
We then report the averaged scores over all $N$ pairs.

\noindent\textbf{Evaluated models.}
To analyze how model characteristics affect verification performance, our model selection in evaluation is guided by three criteria. First, we include models with varying parameter scales to examine the impact of model capacity. Second, for audio-capable models, we evaluate both video-only and video+audio settings to assess the contribution of audio signals to event verification. Third, we evaluate powerful closed-source models, including GPT 5 mini~\cite{singh2025openai}, as strong reference points for current model performance. Detailed model configurations and the evaluation prompt are provided in Appendix~\ref{appendix_exp}.

\begin{figure}
    \centering
    \includegraphics[width=0.95\linewidth]{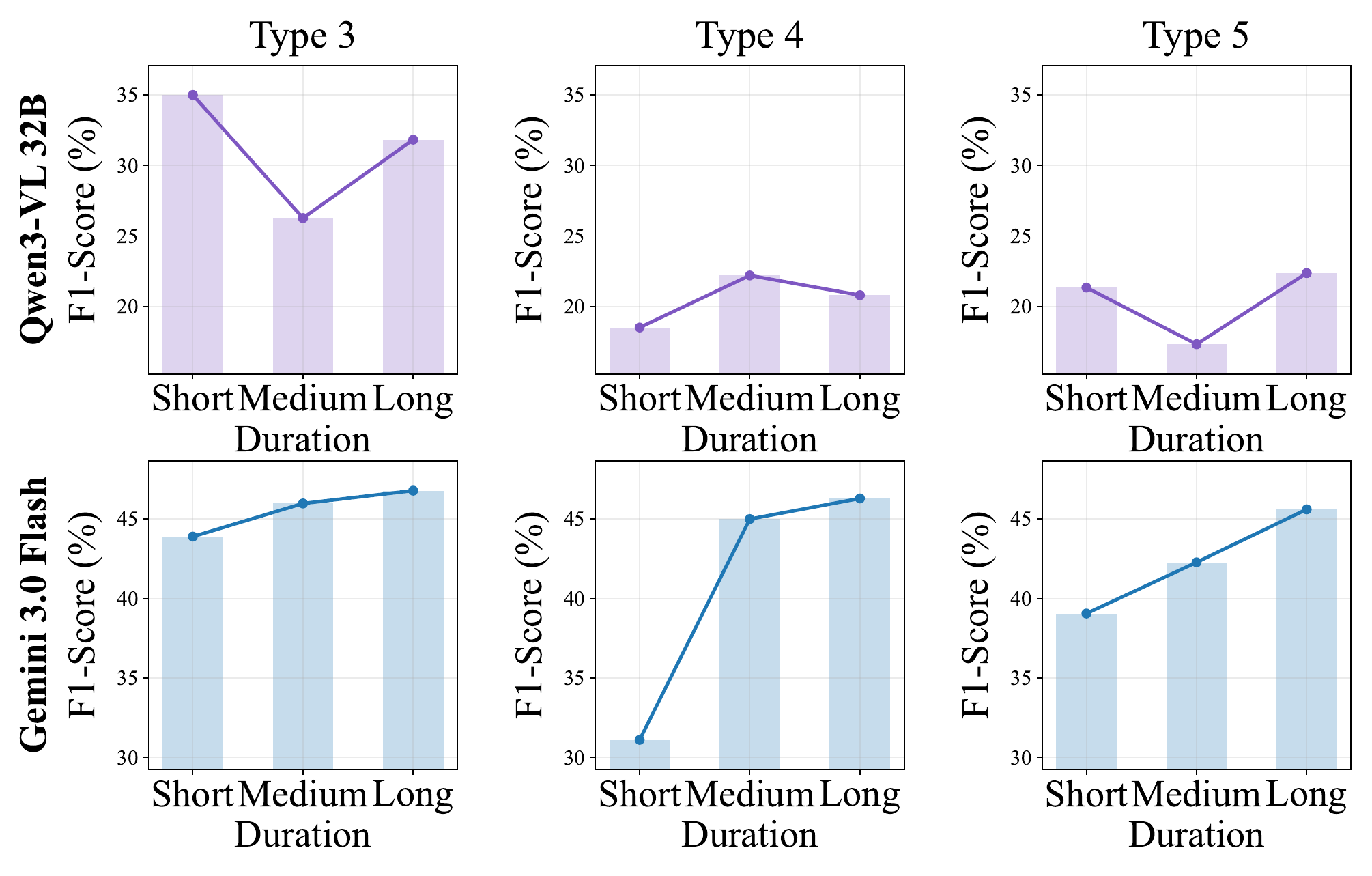}
    \caption{\textbf{F1-Score variations by video duration under fixed complexity.} Each column represents a specific complexity type, illustrating how performance changes across Short, Medium, and Long durations.}
    \label{fig:qwen_gemini_weak}
\end{figure}

\subsection{Overview of Results}
Table~\ref{tab:main_results} presents the quantitative evaluation of diverse Video-LLMs on \dataset, together with their performance on existing benchmarks. Qualitative examples of Video-LLM outputs are shown in Figure~\ref{fig:example_haldet}.

\noindent\textbf{VidOmni-Bench enables rigorous evaluation beyond existing benchmarks.}
Our sentence-wise event verification task effectively differentiates Video-LLM performance. High leaderboard scores on existing benchmarks such as Video-MME~\cite{Fu_2025_CVPR}, Video-MMMU~\cite{hu2025video}, and LVBench~\cite{wang2025lvbench} do not necessarily translate to strong performance on VidOmni-Bench. We further observe a significant gap between Precision and Recall across most models, indicating that they often fail to consistently identify hallucinated sentences. These findings show that \dataset provides a more challenging testbed for fine-grained video understanding than existing evaluation benchmarks.

\noindent\textbf{Performance comparison.} 
Gemini 3 Flash shows the highest overall performance when using both video and audio inputs, serving as the strongest reference point among evaluated settings. Among open-source models, Qwen3-VL 32B performs best, while the InternVL3.5 family also shows competitive results. However, Figure~\ref{fig:label_recall} shows that InternVL3.5 38B is particularly strong at verifying static-frame-related hallucinations but weaker on temporal errors. This suggests that strong frame-level understanding does not necessarily translate to robust temporal reasoning.

\noindent\textbf{Examples of Video-LLMs' output}. Figure~\ref{fig:example_haldet} shows sentence-wise event verification results from multiple Video-LLMs on a Qwen3-VL-generated caption. Each model outputs a sentence-level prediction together with reasoning, allowing us to qualitatively compare not only the final judgments but also the evidence used to support them.

\noindent\textbf{Effect of scaling model size on event verification.} 
Performance generally improves with model scale within the same family. Larger models achieve higher F1-Scores, suggesting that increased parameter capacity enhances their ability to detect subtle discrepancies between hard negative sentences and the corresponding video evidence.

\noindent\textbf{Effect of audio modality on event verification.} 
As shown in Table~\ref{tab:main_results}, adding audio consistently decreases the F1-Scores within the video-SALMONN 2+ family, indicating that audio may introduce cross-modal noise when it is not properly grounded with visual evidence. In contrast, Gemini 3 benefits from audio input, where audio often provides verification-relevant cues complementary to visual information. This suggests that audio can improve sentence-wise event verification when models effectively identify and integrate audio evidence relevant to each caption sentence.

\subsection{Detailed Analysis}

\begin{figure}
    \centering
    \includegraphics[width=0.9\linewidth]{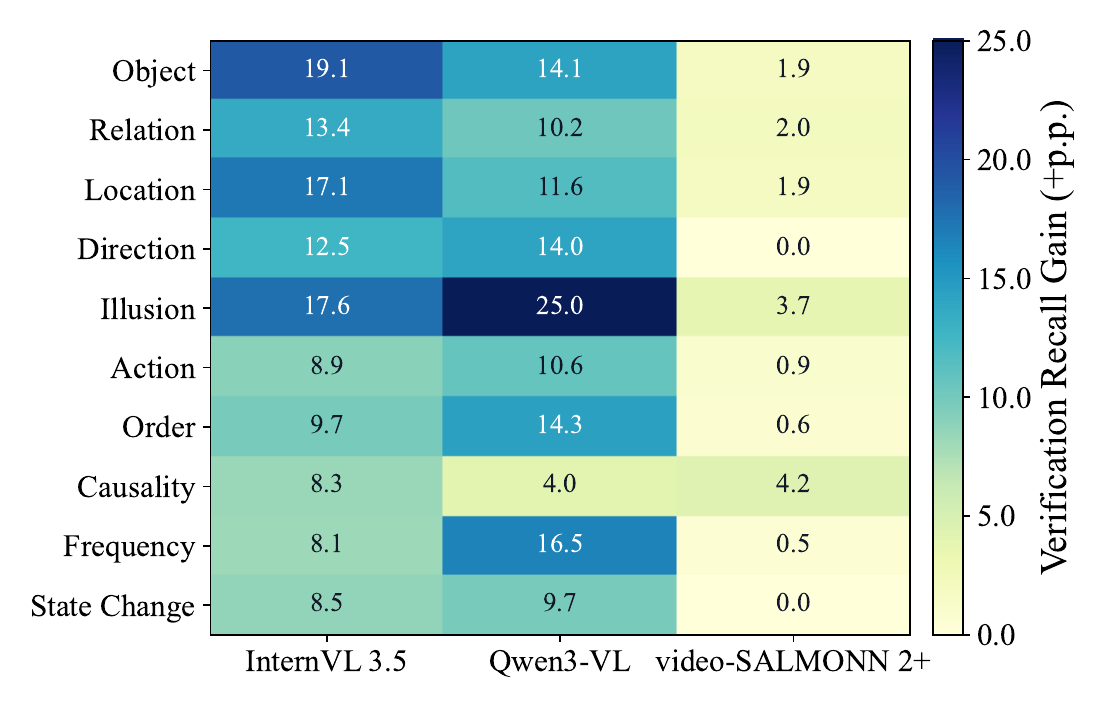}
    \caption{\textbf{Recall gain by error category as model size increases.} The heatmap shows the improvement in Recall for InternVL3.5, Qwen3-VL, and video-SALMONN 2+ when scaling up parameters.}
    \label{fig:label_gain}
\end{figure}

\noindent\textbf{Success cases of event verification.} 
As illustrated in Figure~\ref{fig:example_haldet}, successful verification is often tightly coupled with factually grounded reasoning. This indicates that the model's output is based on an integrated understanding of the video content.

\noindent\textbf{Failure cases of event verification.} 
We observe two primary failure patterns. First, models may predict an Incorrect sentence as Correct by overlooking its mismatch with the video evidence. Second, even correct predictions can be accompanied by incomplete or factually flawed explanations. These cases highlight the need to examine both predictions and explanations when assessing whether verification is grounded in the video evidence.

\noindent\textbf{Video-LLMs have their own weak complexity and duration.} 
Figure~\ref{fig:qwen_gemini_weak} shows that performance trends across complexity and duration differ by model. For example, Gemini 3 Flash exhibits a consistent F1-Score decline as duration shortens within high-complexity types, while Qwen3-VL shows different vulnerabilities depending on the complexity type. These results demonstrate that VidOmni-Bench diagnoses multidimensional, model-specific weaknesses beyond a single difficulty axis. Further analyses are provided in the Appendix~\ref{appendix_exp}.

\noindent\textbf{Category-wise performance gains from model size scaling.} To examine where performance improvements arise as Video-LLMs scale, we measure Recall gains across error categories when increasing model size for InternVL3.5 (2B to 38B), Qwen3-VL (2B to 32B), and video-SALMONN 2+ (7B to 72B). Figure~\ref{fig:label_gain} shows that scaling consistently improves Recall more for static-frame-related errors than for temporal-related errors. These results suggest that increasing model size within the same family mainly improves verification of static visual attributes and spatial relations, whereas temporal errors remain challenging.

\begin{figure}
    \centering
    \includegraphics[width=1.0\linewidth]{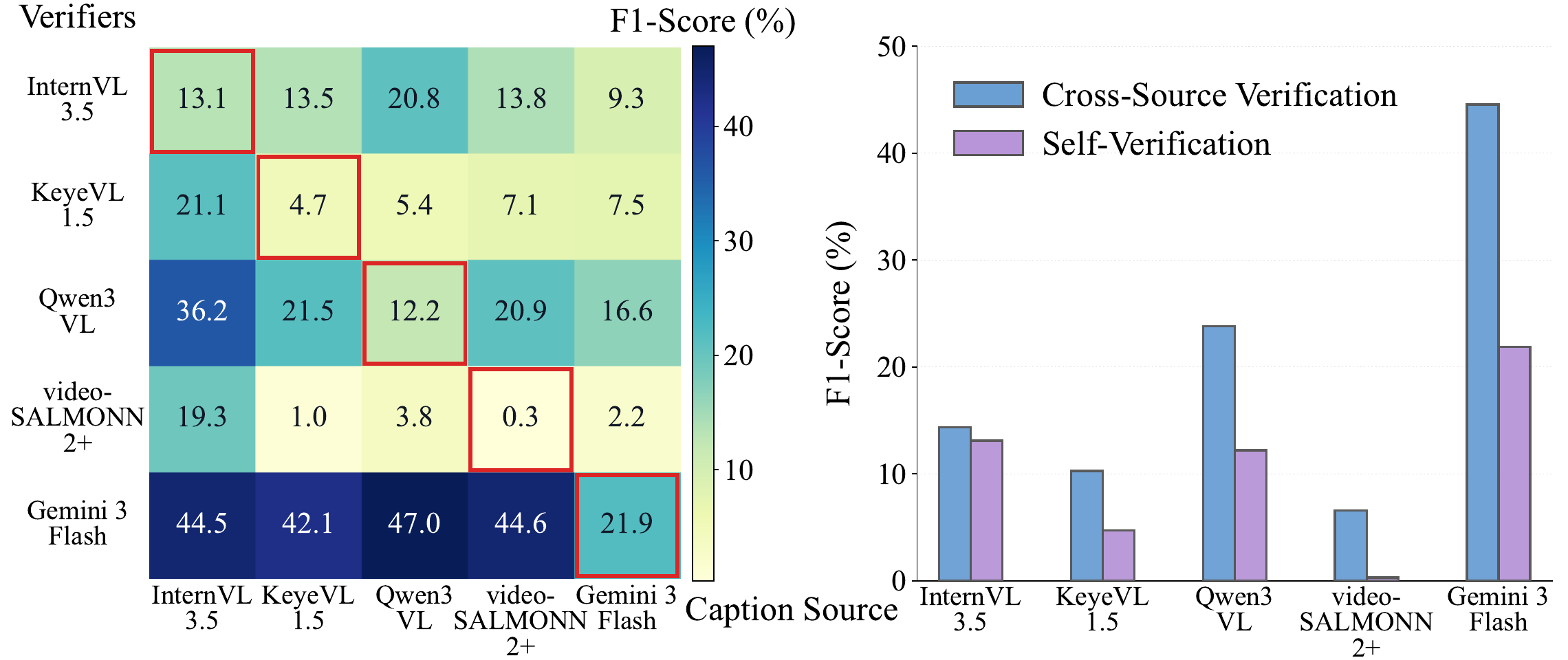}
    \caption{\textbf{Self-preference bias in self-verification.} \textbf{Left:} The heatmap shows F1-Scores for sentence-wise verification, with the y-axis representing the evaluated models and the x-axis representing the caption source models. \textbf{Right:} The bar chart compares the performance on cross-source verification versus self-verification.}
    \label{fig:self_bias}
\end{figure}

\noindent\textbf{Video-LLMs struggle to detect their own hallucinations}.
As shown in Figure~\ref{fig:self_bias}, Video-LLMs perform worse on their own captions than on captions generated by other models. This trend is clear even for Gemini 3 Flash, the best-performing model. Its F1-Score exceeds 40 on captions from other models but drops by nearly half on its own captions. This suggests a strong self-preference bias, where models tend to judge their own generated sentences as Correct. Further analysis in the Appendix~\ref{appendix_exp} shows that ensemble-based verification can mitigate this bias. These results indicate that self-evaluation remains a significant challenge for current Video-LLMs.

\noindent\textbf{Additional analysis in the Appendix}.
The Appendix~\ref{appendix_exp} provides additional robustness analyses for \dataset. Increasing the input frames does not yield clear performance improvements, indicating that the benchmark's difficulty cannot be resolved simply by sampling more frames. We also evaluate a subset excluding YouTube-sourced videos and observe similar trends, suggesting that the strong performance of Gemini 3 and GPT 5 is unlikely to be driven primarily by data leakage.

\section{Conclusion}
We introduced \dataset, a sentence-wise event verification benchmark for fine-grained video understanding in Video-LLMs. By requiring models to verify dense caption sentences across video complexity and duration, \dataset reveals model-specific vulnerabilities and self-preference bias. These findings show that even strong models still struggle with reliable, temporally grounded event verification. We hope \dataset serves as a rigorous evaluation standard for more faithful fine-grained video understanding.

\section*{Limitations}
VidOmni-Bench has several limitations that suggest directions for future work. First, verifying dense video captions inevitably involves some ambiguity, especially for complex temporal events. We mitigate this issue by introducing an Unknown category and using a multi-step labeling protocol, but complete agreement can still be difficult in borderline cases. Second, our analysis reveals self-preference bias in Video-LLMs, suggesting that model-based caption generation and evaluation may benefit from more diverse caption sources and cross-verification protocols. Third, the benchmark construction depends on the generative capability of current Video-LLMs and requires human verification to ensure data quality, which may limit scalability. Despite these limitations, VidOmni-Bench provides a useful diagnostic resource for studying current gaps in fine-grained video understanding.

\section*{Ethics Statement}
\noindent\textbf{AI assistant usage in writing.} ChatGPT \citep{singh2025openai} was used to improve readability during manuscript preparation.
The authors reviewed and edited all generated text and take full responsibility for the final content.

\noindent\textbf{Dataset construction.} \dataset is constructed from 
publicly available video datasets together with supplementary videos 
crawled from YouTube under Creative Commons licenses. To respect the 
copyright and terms of service of the original content providers, we 
do not redistribute raw video files for any of the 500 videos in 
\dataset, regardless of source. Instead, we will release only video metadata, the generated dense captions, and the annotations upon publication. We note that \dataset includes 
videos from anomaly detection datasets that 
may depict violent or disturbing events, as well as footage that may 
contain identifiable individuals. We apply no additional filtering 
beyond the curation already performed by the original providers, and 
recommend that users exercise discretion accordingly.

\section*{Acknowledgements} This research was supported by the Institute of Information \&communications Technology Planning \& Evaluation (IITP) grant funded by the Korea government(MSIT) (No. RS-2019-II190079, Artificial Intelligence Graduate School Program(Korea University), 1\%; No. RS-2025-25439490, 40\%), the KOCCA grant (RS-2024-00345025, 10\%), the National Research Foundation of Korea(NRF) grant funded by the Korea government(MSIT)(No. RS-2025-02263628, 34\%) This research was supported by JST PRESTO, Japan, Grant Number JPMJPR2523 (5\%). This work was partly achieved through the use of SQUID at D3 Center, The University of Osaka (5\%). This work was supported in part by the Physical AI Development Support Program by AWS Japan through the provision of computational resources (5\%).

\bibliography{custom}

\appendix
\clearpage

\section*{Appendix}

\section{Attribution of Icons}
The icons used in the figures are sourced from Flaticon (\url{https://www.flaticon.com/}). We gratefully acknowledge the creators for providing these visual assets, which are utilized in accordance with their licensing terms to enhance the clarity of our technical illustrations.

\section{Construction of \dataset}
\label{appendix_dataset}

\subsection{Video Collection}

\noindent\textbf{More details of video collection.}
Before collecting videos for \dataset, we first define a taxonomy that categorizes videos according to the complexity of their spatio-temporal dynamics. The taxonomy consists of five complexity types, where each type progressively introduces an additional source of difficulty: motion, fine-grained action, multi-agent binding, and scene transitions.

This design results in five complexity types. Static World videos contain little or no temporal change, so they can largely be understood from static visual content. Body Motion videos introduce temporal changes caused by human or object motion, such as walking, running, or turning. Hand-Object Interaction videos further require understanding fine-grained local actions, such as grasping, placing, or manipulating an object. Crowd Action videos involve multiple agents acting simultaneously, where the model must correctly associate each action with the corresponding subject or object. Finally, Multi-Scene Dynamics videos contain shot changes or multiple scenes, requiring the model to connect events across temporally discontinuous segments.

We note that these types are not intended to be mutually exclusive definitions of intrinsic video complexity. Instead, our taxonomy is designed as a controlled diagnostic dimension, where each type emphasizes a different source of spatio-temporal difficulty. As shown in Figure~\ref{fig:aggregated_result}, this taxonomy provides a useful axis for examining how Video-LLMs behave under different video dynamics. Tables~\ref{tab:appendix_complexity_taxonomy} and~\ref{tab:appendix_video_complexity} summarize the complexity taxonomy and the public video sources used for each type.

\begin{center}
\renewcommand{\arraystretch}{1.2}
\resizebox{\columnwidth}{!}{%
\begin{tabular}{l c c c c}
\toprule
\textbf{Complexity} 
& \textbf{Motion} 
& \makecell{\textbf{Fine-grained}\\\textbf{Action}} 
& \makecell{\textbf{Multi-Agent}\\\textbf{Binding}} 
& \makecell{\textbf{Scene}\\\textbf{Transitions}} \\
\midrule
Type 1 & \xmark & \xmark & \xmark & \xmark \\
Type 2 & \cmark & \xmark & \xmark & \xmark \\
Type 3 & \cmark & \cmark & \xmark & \xmark \\
Type 4 & \cmark & \cmark & \cmark & \xmark \\
Type 5 & \cmark & \cmark & \cmark & \cmark \\
\bottomrule
\end{tabular}
}
\captionof{table}{\textbf{Complexity taxonomy of VidOmni-Bench.}}
\label{tab:appendix_complexity_taxonomy}
\end{center}

\begin{table}[ht]
\centering
\renewcommand{\arraystretch}{1.15}
\resizebox{\columnwidth}{!}{%
\begin{tabular}{c c l}
\toprule
\textbf{Complexity} 
& \textbf{Name} 
& \textbf{\makecell{Video Source\\(Public Dataset)}} \\
\midrule
Type 1 
& Static World 
& \makecell[l]{Video-ACID~\cite{8734060}\\ SpatialVID~\cite{wang2025spatialvid}\\ Mannequin Challenge~\cite{li2020mannequinchallenge}} \\
\midrule
Type 2 
& Body Motion 
& \makecell[l]{ActivityNet~\cite{caba2015activitynet}} \\
\midrule
Type 3 
& Hand-Object Interaction 
& \makecell[l]{Assembly101~\cite{sener2022assembly101}\\ Epic-Kitchens~\cite{damen2018scaling}\\ Ego4D~\cite{grauman2022ego4d}} \\
\midrule
Type 4 
& Crowd Action 
& \makecell[l]{UCF-Crime~\cite{Sultani_2018_CVPR}\\ Sports-1M~\cite{Karpathy_2014_CVPR}} \\
\midrule
Type 5 
& Multi-Scene Dynamics 
& \makecell[l]{Video-MME~\cite{Fu_2025_CVPR}\\ XD-Violence~\cite{Wu2020not}} \\
\bottomrule
\end{tabular}}
\caption{\textbf{Data sources of each complexity type in \dataset}. Videos are primarily collected from public datasets. We further supplement underrepresented types and duration ranges with videos sourced from YouTube.}
\label{tab:appendix_video_complexity}
\end{table}

\subsection{Dense Video Caption Generation}

\noindent\textbf{Captioning setups.} Table~\ref{tab:video_captioner_details} summarizes the Video-LLMs used for caption generation. Open-source models are deployed with official implementations and publicly available weights. Input Frames Per Second (FPS) and the maximum number of input frames are set following reported configurations from Video-MME~\cite{Fu_2025_CVPR}, Video-MMMU~\cite{hu2025video}, and LVBench~\cite{wang2025lvbench} whenever available. Otherwise, we carefully set configurations that cover as much video content as possible within each model's context length. We fix the generation seed for reproducibility. For the Video-LLMs used for dense captioning, we randomly select one of the instructions in Figure~\ref{fig:captioning_prompt}.

\begin{table}[h]
\centering
\renewcommand{\arraystretch}{1.2}
\resizebox{\columnwidth}{!}{%
\begin{tabular}{l c c c c c}
\toprule
\textbf{Model} & \textbf{Provider} & \textbf{Open/Closed} & \textbf{FPS} & \textbf{Max Frames} & \textbf{Release} \\ \midrule
Gemini 3 Flash & Google & Closed & 1 & - & 2026/02 \\
Qwen3-VL 32B & Alibaba & Open & 2 & 2048 & 2026/01 \\
InternVL 3.5 38B & OpenGVLab & Open & 1 & 32 & 2025/12 \\
video-SALMONN 2+ 72B & ByteDance & Open & 1 & 768 & 2025/11 \\
Keye-VL 1.5 8B & Kwai & Open & 1 & 1024 & 2025/03 \\ \bottomrule
\end{tabular}
}
\caption{\textbf{Details of Video-LLMs chosen as captioning engines.} We use five Video-LLMs with diverse providers, accessibility, FPS, and maximum number of frames (Max Frames) to generate dense captions.}
\label{tab:video_captioner_details}
\end{table}

\begin{figure}[t]
\small
\centering
\begin{promptbox}{Instruction given to Video-LLMs in captioning}
\begin{enumerate}[leftmargin=1.5em, itemsep=0.55em, topsep=0.2em]
    \item Describe this video in detail, focusing on the spatio-temporal dynamics. Describe exactly how objects and agents move, change, and occupy space over time within the scene.

    \item Give a detailed account of everything shown in the video, capturing all visible specifics. Describe events in the exact order they appear over time. Ensure that you describe the sequence of events exactly as they occur, without skipping any steps.

    \item Describe the video in detail, paying special attention to how objects and people interact with each other. Capture the precise timing and nature of every contact, movement, and reaction shown in the footage.

    \item Thoroughly describe the video’s visual narrative, capturing every visible detail from start to end. Emphasize how actions unfold and how the scene transitions logically over time.

    \item Provide a thorough description of every detail, explicitly prioritizing spatio-temporal dynamics. Capture all visual elements, and focus on their spatial positions, movement trajectories, and how the scene evolves over time.
\end{enumerate}
\end{promptbox}
\caption{\textbf{Prompt templates for dense video captioning.} We use five prompt templates to generate dense captions for each video, reducing the potential instruction bias that may arise from relying on a single captioning prompt.}
\label{fig:captioning_prompt}
\end{figure}

\subsection{Sentence-wise Annotation}

\noindent\textbf{Annotator.} 
We recruit five annotators specifically for our multi-stage labeling task, which requires meticulous reading and temporal reasoning to identify hallucinated sentences within dense captions. All candidates are required to demonstrate high-level English proficiency and the ability to detect subtle spatio-temporal inconsistencies. These qualifications are rigorously verified through a pilot annotation task involving 10 videos prior to the main annotation phase.

\noindent\textbf{Voting and quality control.}
The authors review all pilot annotations, and only those who demonstrated high accuracy in identifying hallucinated sentences are designated as trusted annotators for the main task. Each trusted annotator is assigned between 500 and 1,000 video-caption pairs. To ensure objective and robust annotation, each pair is independently evaluated by at least two annotators, including a minimum of one trusted annotator. Once at least two annotations are collected for a sentence, the final label is determined through majority voting. Furthermore, the authors conduct systematic audits for every batch of 100 video-caption pairs. If a batch fails to meet our criteria, annotators are required to re-evaluate the entire set before proceeding.

\noindent\textbf{Inter-annotator agreement.}
We evaluate inter-annotator agreement on VidOmni-Bench using Fleiss’ kappa \cite{fleiss1971measuring} with labels (Correct, Incorrect, Unknown). The overall agreement reaches $\kappa$ = 0.50 with an observed agreement of $P_{bar}$ = 0.764. While these values indicate a moderate level of agreement, we prioritize data integrity through a rigorous manual adjudication process. To ensure the absolute reliability of the benchmark, authors meticulously review and resolve all disputed labels and cases with low confidence, establishing a high-quality ground truth for subsequent evaluation.

\begin{figure*}
    \centering
    \includegraphics[width=1.0\linewidth]{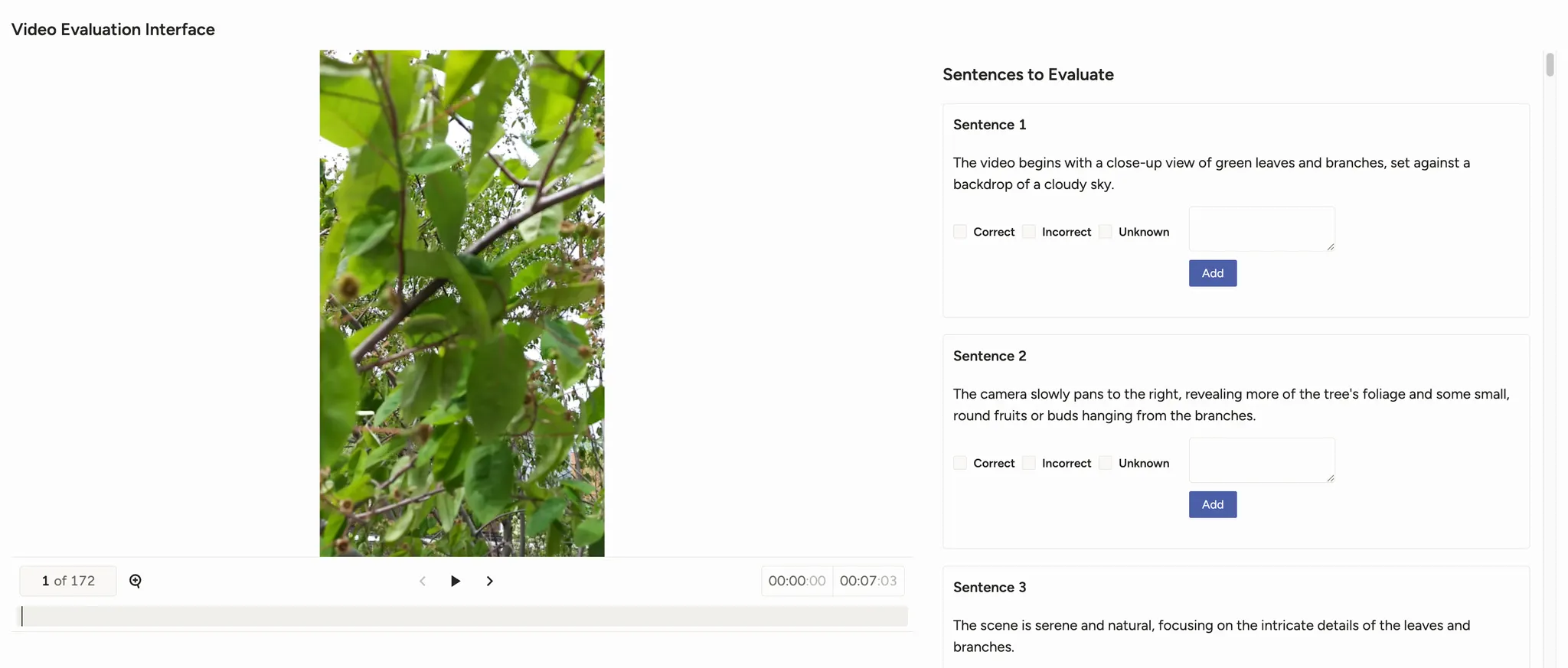}
    \caption{\textbf{Annotation interface for sentence-wise correctness (Stage 1)}. Annotators label each caption sentence as Correct, Incorrect, or Unknown based on the video evidence. An additional memo field is provided to collect detailed notes and ambiguous cases during dataset construction.}
    \label{fig:step1}
\end{figure*}

\begin{figure*}
    \centering
    \includegraphics[width=1.0\linewidth]{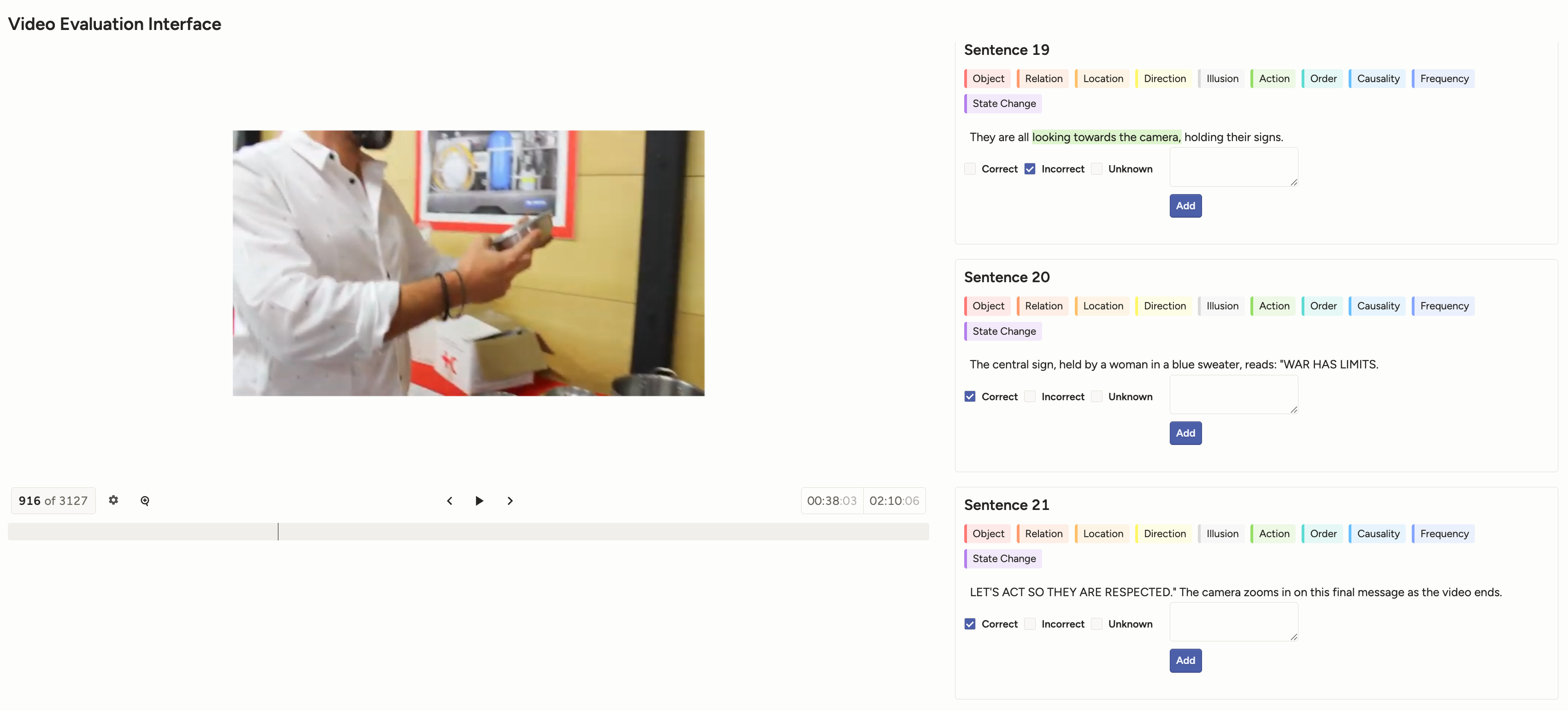}
    \caption{\textbf{Annotation interface for word-level error category (Stage 2).} For each sentence labeled as Incorrect in Stage 1, annotators identify the erroneous words or phrases and assign them to predefined error categories, enabling fine-grained analysis of hallucination types.}
    \label{fig:step2}
\end{figure*}

\noindent\textbf{Annotation interface.}
We customize the Label Studio~\cite{LabelStudio} interface for our two-stage annotation pipeline. Figure~\ref{fig:step1} shows the interface for sentence-wise correctness annotation (Stage 1), while Figure~\ref{fig:step2} shows the interface for word-level error category annotation (Stage 2).

\noindent\textbf{Word-level error categories.}
Table \ref{tab:error_types} provides detailed descriptions for each error categories used during our annotation process. For every sentence labeled as Incorrect, annotators are instructed to pinpoint the specific errors at the word level and classify them according to these categories.

\begin{table*}[ht]
\centering
\small
\begin{tabular}{lp{5.5cm}p{6.5cm}}
\toprule
\textbf{Type} & \textbf{Description} & \textbf{Example} \\ \midrule
Object & Misdescriptions of the object itself including its attributes (color, text, count, etc.). & There are two blue cars, but described as ``three red cars.'' \\
Relation & Misdescriptions of spatial relationships between objects. & A cup is under the table, but described as ``on the table.'' \\
Location & Misdescriptions of an object's position. & A cup is on the right side, but described as ``on the left.'' \\
Direction & Misdescriptions of orientation or direction of an object. & A person is looking to the right, but described as ``looking to the left.'' \\
Illusion & Mentioning non-existent objects, events, or actions. & No bird is present, but described as ``a bird flying in the sky.'' \\
Action & Incorrect or incomplete description of actions over time. & A person is walking, but described as ``running.'' \\
Order & Misdescriptions of the temporal order of events or actions. & Sat down after opening a door, but described as ``door opened after sitting down.'' \\
Causality & Misdescriptions of causal relationships among events. & Describing that rain was caused by people opening umbrellas. \\
Frequency & Incorrect tracking of the frequency of repeated actions. & A person jumped once, but described as ``jumped three times.'' \\
State Change & Misdescribing an object's state transition. & Traffic light changed from red to yellow, but described as ``yellow to red.'' \\ \bottomrule
\end{tabular}
\caption{\textbf{Word-level error categories in \dataset.} We group word-level errors into five static-frame-related categories(Object, Relation, Location, Direction, and Illusion) and five temporal-understanding-related categories(Action, Order, Causality, Frequency, and State Change).}
\label{tab:error_types}
\end{table*}

\subsection{Edge Case Handling Guidelines}
Due to the length and complexity of generated captions, we establish internal guidelines to maintain consistency, particularly for ambiguous or repetitive sentences that are difficult to classify using our error categories. These guidelines function as a decision rules to help annotators navigate edge cases systematically.

\noindent\textbf{Handling repetitive sentences.}
A common failure mode in Keye-VL 1.5~\cite{yang2025kwai} and video-SALMONN 2+~\cite{tang2025videob} is that a model repeatedly generates the same or similar sentences. In such cases, we evaluate the first occurrence based on its correctness. If the video progresses to a new event but the model continues to repeat the previous sentence, all subsequent repetitions are labeled Incorrect and assigned the Frequency error. This ensures the annotation reflects the model's inability to track temporal progression.

\noindent\textbf{Distinguishing temporal errors.}
We provide explicit instructions to help annotators distinguish events that are temporally misplaced from those that are entirely hallucinated. For example, if a model describes an event that does not occur at the specified timestamp but happens elsewhere in the video, it is labeled as an Order error rather than an Illusion error. This distinction is crucial for diagnosing whether a model suffers from total hallucination or merely temporal shuffling of frames.

\noindent\textbf{Resolving visual ambiguity.}
In cases where the visual evidence is too ambiguous to make a definitive judgment, annotators utilize the Unknown category or seek final adjudication through discussions with the authors. This collaborative feedback loop ensures that the final dataset remains a reliable ground truth for fine-grained video understanding.

\section{Details of Experimental Setups}
\label{appendix_exp}

\subsection{Details of Evaluation on VidOmni-Bench}
\noindent\textbf{Source of models.}
For open-weight models, we use checkpoints available on Hugging Face~\cite{wolf2020transformers} and base our code on the Hugging Face Transformers package. For GPT 5 mini and Gemini 3 Flash, we use their official APIs.

\noindent\textbf{Prompt for sentence-wise event verification.} 
To ensure a fair and consistent evaluation, we apply a unified chat template across all evaluated Video-LLMs. The full prompt template is shown in Figure~\ref{fig:verification_prompt}. For each inference, the prompt remains identical except for the specific target caption provided for sentence-wise verification.

\begin{figure}[t]
\small
\centering
\begin{promptbox}{Instruction given to Video-LLMs in sentence-wise verification}
You are an expert in Video Hallucination Detection. You are given a video and a caption that consists of multiple sentences. Your task is to evaluate each sentence individually to determine if it contains a hallucination based on the visual evidence.

\vspace{0.6em}
For each sentence, assign a binary label (0 or 1) according to the following definition:

\vspace{0.6em}
- Label 1: The sentence contains a hallucination. This means it describes something not present or incorrect regarding the video.

\vspace{0.4em}
- Label 0: The sentence is faithful to the video. This means it correctly describes the visual content.

\vspace{0.6em}
Please first output the dictionary of labels before explaining the reason for your judgment.

\vspace{0.6em}
Output format requirements:

\vspace{0.3em}
1) First line must be a valid JSON dictionary with keys sentence\_1, sentence\_2, ... and values 0 or 1.

\vspace{0.3em}
2) Then explain the reason for each sentence.

\vspace{0.6em}
Caption:
\end{promptbox}
\caption{\textbf{Sentence-wise event verification instruction.} The prompt asks Video-LLMs to assign a binary label to each caption sentence before providing sentence-level reasoning.}
\label{fig:verification_prompt}
\end{figure}

\noindent\textbf{Computation.}
At most four H200 GPUs are used for inference of a single model.

\noindent\textbf
{Evaluation configs.}
As in our captioning pipeline, we set FPS and the maximum number of input frames following each benchmark's reported settings whenever available, and otherwise sample as many frames as possible up to each model's native context limit, as detailed in Table~\ref{tab:model_config_details}.

\begin{table}[h]
\centering
\small
\renewcommand{\arraystretch}{1.2}
\resizebox{\columnwidth}{!}{%
\begin{tabular}{l c c c c c}
\toprule
\textbf{Model} & \textbf{Provider} & \textbf{Open/Closed} & \textbf{FPS} & \textbf{Max Frames} & \textbf{Release} \\ \midrule
Gemini 3 Flash & Google & Closed & 1 & - & 2026/02 \\
GPT 5 mini & OpenAI & Closed & - & 256 & 2026/03 \\
Qwen3-VL 32B & Alibaba & Open & 2 & 2048 & 2026/01 \\
InternVL 3.5 38B & OpenGVLab & Open & 1 & 768 & 2025/12 \\
Video-SALMONN 2+ 72B & ByteDance & Open & 1 & 32 & 2025/11 \\
Keye-VL 1.5 8B & Kwai & Open & 1 & 1024 & 2025/03 \\ \bottomrule
\end{tabular}
}
\caption{\textbf{Details of Video-LLMs evaluated on \dataset.} We list the provider, accessibility, FPS, and maximum number of input frames (Max Frames) for each Video-LLM evaluated on \dataset.}
\label{tab:model_config_details}
\end{table}

\section{Additional Analysis}
\label{additional_analysis}

\subsection{Caption Analysis in \dataset}

\noindent\textbf{Details for Certificate Coverage.}
We follow Video-MME~\cite{Fu_2025_CVPR}, which defines Certificate Length as the minimum video span required to solve a task. For sentence-wise event verification, we annotate the minimum video intervals needed to determine whether each sentence is supported by the video. Certificate Coverage is then computed as the ratio of the annotated Certificate Length to the full video duration. Since annotating Certificate Length is highly labor-intensive, the Certificate Coverage reported in Figure~\ref{fig:vidomni_validity} and Table~\ref{tab:benchmark_comparison} is computed on a randomly sampled subset of 30 videos for each benchmark.

\noindent\textbf{Captions of \dataset are more complex than other benchmarks.}
We compare the informativeness of captions in \dataset with those from ARGUS~\cite{rawal2025argus}, NOAH~\cite{lee2025noah}, and CAPAbility~\cite{liu2025capability}. We first conduct a lexical and syntactic analysis using spaCy~\cite{Honnibal_spaCy_Industrial-strength_Natural_2020}. As shown in Table~\ref{tab:caption_informativeness}, captions in \dataset contain substantially more words, information units, noun chunks, and verbs than captions from prior benchmarks.

In addition to these surface-level statistics, we further assess caption informativeness through a pairwise LLM-as-a-judge evaluation following Prometheus-2~\cite{kim2024prometheus}. Given the predefined informativeness rubric and prompt shown in Figure~\ref{fig:informativeness_prompt}, the judge compares each \dataset caption with a caption from another benchmark and selects the more informative one. \dataset captions win in 96.0--99.6\% of pairwise comparisons, suggesting that they provide richer and more detailed event descriptions for sentence-wise event verification.

\setcounter{table}{9}
\begin{table*}[t]
\centering
\small
\renewcommand{\arraystretch}{1.15}
\resizebox{0.9\textwidth}{!}{%
\begin{tabular}{lccccc}
\toprule
\textbf{Benchmark}
& \textbf{\makecell{Avg. \\ Words}}
& \textbf{\makecell{Info \\ Count}}
& \textbf{\makecell{Noun \\ Chunks}}
& \textbf{Verbs}
& \textbf{\makecell{LLM Judge Win Rate (\%) \\ (vs. Ours) }} \\
\midrule
ARGUS~\cite{rawal2025argus}
& 477.5
& 336.4
& 134.3
& 63.1
& 4.0 \\
NOAH~\cite{lee2025noah}
& 48.0
& 34.2
& 13.8
& 8.0
& 0.6 \\
CAPAbility~\cite{liu2025capability}
& 3.2
& 2.4
& 1.1
& 0.3
& 0.4 \\
\midrule
\dataset
& 578.9
& 400.6
& 159.3
& 82.1
& -- \\
\bottomrule
\end{tabular}
}
\caption{\textbf{Comparison of caption informativeness across benchmarks.}
\dataset achieves substantially longer captions, higher information counts, and more noun chunks and verbs than prior benchmarks. The last column reports the percentage of pairwise LLM-as-a-judge comparisons in which the baseline caption is preferred over the \dataset caption in terms of informativeness.}
\label{tab:caption_informativeness}
\end{table*}

\begin{figure*}[t]
\small
\centering
\begin{promptbox}{Prompt for pairwise caption informativeness judgment}
\textbf{Task Description:}

\vspace{0.3em}
Compare Response A and Response B using the rubric. Write one short feedback sentence, then end with exactly one result tag. The final line must be exactly either:

\vspace{0.3em}
\texttt{[RESULT] A}

\texttt{[RESULT] B}

\vspace{0.6em}
\textbf{Instruction:}

\vspace{0.3em}
Describe this video in detail, focusing on the spatio-temporal dynamics. Describe exactly how objects and agents move, change, and occupy space over time within the scene.

\vspace{0.6em}
\textbf{Response A:}

\vspace{0.3em}
\texttt{\{response\_A\}}

\vspace{0.6em}
\textbf{Response B:}

\vspace{0.3em}
\texttt{\{response\_B\}}

\vspace{0.6em}
\textbf{Rubric:}

\vspace{0.3em}
Does the model's response provide a sufficient amount of relevant and substantive information for the given query?

\vspace{0.4em}
Score 1: The response is severely lacking in information. It is vague, overly generic, or omits essential details, leaving the query largely unaddressed.

\vspace{0.3em}
Score 2: The response provides minimal information. It touches on the topic but lacks depth, with significant gaps or important aspects left uncovered.

\vspace{0.3em}
Score 3: The response provides a moderate amount of information. It covers the core of the query but may lack depth in some areas or omit secondary details that would add value.

\vspace{0.3em}
Score 4: The response is informative and substantive. It thoroughly addresses the query with relevant details, with only minor opportunities for additional depth or coverage.

\vspace{0.3em}
Score 5: The response is highly informative and comprehensive. It fully addresses the query with rich, relevant, and well-elaborated detail, leaving little need for follow-up.

\vspace{0.6em}
\textbf{Feedback:}
\end{promptbox}
\caption{\textbf{Prompt for pairwise caption informativeness judgment.}
We use Prometheus-2 7B to compare captions from \dataset and other benchmarks under a predefined informativeness rubric. Given the captioning instruction, two captions, and the rubric, the judge selects the more informative caption.}
\label{fig:informativeness_prompt}
\end{figure*}

\subsection{Additional Experiments on \dataset}

\noindent\textbf{\dataset remains challenging beyond frame sampling.}
Table~\ref{tab:frame_sampling_rate} reports type-wise F1-Scores of Qwen3-VL 32B under different FPS from 1 to 10. Overall, F1-Scores show no consistent improvement as FPS increases, indicating that poor performances in Table~\ref{tab:main_results} are not primarily driven by insufficient frame sampling.

\begin{center}
\small
\renewcommand{\arraystretch}{1.15}
\resizebox{\columnwidth}{!}{%
\begin{tabular}{l c c c c c c}
\toprule
\textbf{Video-LLM} & \textbf{FPS} & \textbf{Type 1} & \textbf{Type 2} & \textbf{Type 3} & \textbf{Type 4} & \textbf{Type 5} \\
\midrule
\multirow{5}{*}{Qwen3-VL 32B}
& 1  & 27.32 & 16.21 & 22.18 & 37.08 & 25.22 \\
& 2  & 15.19 & 14.44 & 10.25 & 33.36 & 21.22 \\
& 4  & 18.38 & 15.99 & 31.75 & 37.71 & 22.24 \\
& 8  & 18.38 & 20.09 & 20.14 & 29.04 & 24.63 \\
& 10 & 18.38 & 23.06 & 19.74 & 28.49 & 24.02 \\
\bottomrule
\end{tabular}
}
\setcounter{table}{8}
\captionof{table}{\textbf{Effect of frame sampling rate on Qwen3-VL 32B performance across complexity types.}}
\label{tab:frame_sampling_rate}
\end{center}

\setcounter{table}{10}
\begin{table*}[ht]
\centering
\resizebox{0.85\textwidth}{!}{
\small
\begin{tabular}{llcc}
\toprule
\multirow{2}{*}{\textbf{Video-LLM}} & \multirow{2}{*}{\textbf{Ensemble Verifier}} & \multicolumn{2}{c}{\textbf{F1-Score (\%)}} \\ \cmidrule(lr){3-4}
 & & \textbf{Self-Verification} & \textbf{Cross-Ensemble (+p.p)} \\ \midrule
Qwen3-VL 32B & + InternVL3.5 38B & 5.3 & 26.2 \textbf{(\textcolor{OliveGreen}{+20.9})} \\
InternVL3.5 38B & + Keye-VL 1.5 8B & 7.7 & 24.6 \textbf{(\textcolor{OliveGreen}{+16.9})} \\
Gemini 3 Flash & + GPT 5 mini & 21.9 & 25.1 \textbf{(\textcolor{OliveGreen}{+3.2})} \\ \bottomrule
\end{tabular}
}
\caption{\textbf{Effect of cross-model ensemble on self-verification.} Cross-model ensemble verification consistently improves F1-Score over self-verification across caption sources. These gains show that combining judgments from other Video-LLMs can compensate for the low verification performance observed when models evaluate their own captions.}
\label{tab:cross_model_verification}
\end{table*}

\begin{figure*}
    \centering
    \includegraphics[width=1.0\linewidth]{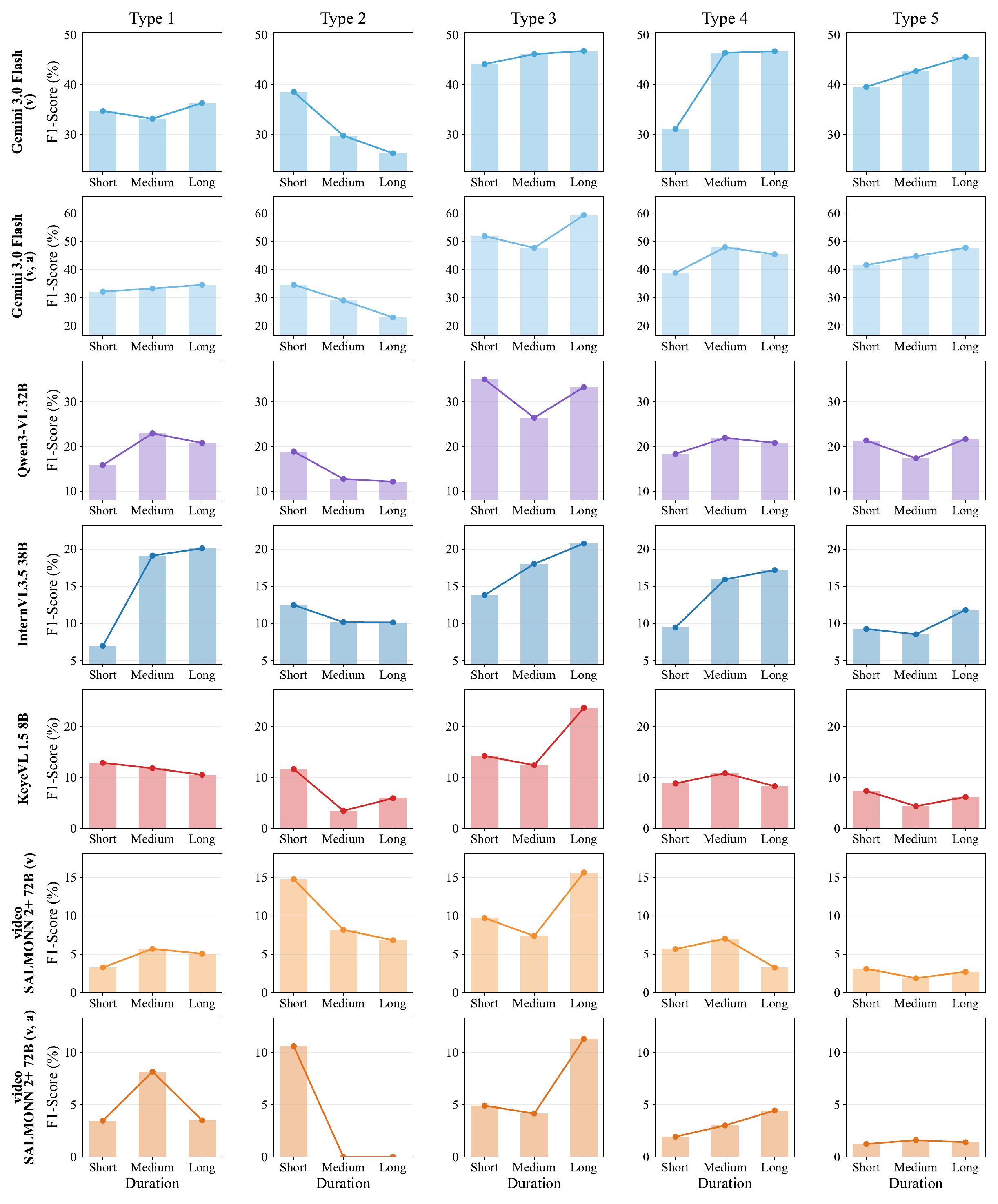}
    \caption{\textbf{Performances of Video-LLMs across all complexity types and durations}. The results reveal multidimensional and model-specific weaknesses, indicating that current Video-LLMs exhibit different failure patterns depending on both video dynamics and temporal scale.}
    \label{fig:aggregated_result}
\end{figure*}

\noindent\textbf{Aggregated performance across all complexity levels and durations.} 
Figure~\ref{fig:aggregated_result} reports the F1-Score of each Video-LLM across the complexity and duration axes of \dataset. The results show that Video-LLM performance varies substantially across different combinations of video dynamics and temporal scale. This reveals multidimensional  weaknesses, where models do not fail uniformly but instead exhibit different vulnerabilities depending on the required visual reasoning and temporal coverage.

\noindent\textbf{YouTube leakage analysis.}
Table~\ref{tab:youtube_leakage} compares model performance on the full VidOmni-Bench test set and a subset excluding YouTube-sourced videos. F1-Scores remain consistent across the two settings, indicating that the results in Table~\ref{tab:main_results} are not substantially driven by YouTube-sourced videos. This suggests that potential YouTube data leakage or source-specific bias has limited impact on the reported performances.

\noindent\textbf{Ensemble can mitigate self-preference bias in self-evaluation.} 
As shown in Table~\ref{tab:cross_model_verification}, cross-model ensemble substantially improves F1-Scores under self-evaluation settings. In this strategy, a generated sentence is marked as Incorrect if either the captioning model or an external verifier predicts it as Incorrect. The consistent gains suggest that self-evaluation failures are largely caused by missed hallucinations, and that external model verification can partially mitigate this self-bias.

\begin{center}
\small
\renewcommand{\arraystretch}{1.2}
\resizebox{\columnwidth}{!}{%
\begin{tabular}{l l c}
\toprule
\textbf{VidOmni-Bench Subset} & \textbf{Video-LLM} & \textbf{F1-Score (\%)} \\
\midrule
\multirow{3}{*}{Full Test Set}
& Gemini 3 Flash & 41.69 \\
& GPT 5 mini & 28.93 \\
& Qwen3-VL 32B & 21.74 \\
\midrule
\multirow{3}{*}{w/o YouTube Videos}
& Gemini 3 Flash & 42.46 \\
& GPT 5 mini & 29.15 \\
& Qwen3-VL 32B & 23.21 \\
\bottomrule
\end{tabular}
}
\captionof{table}{\textbf{YouTube data leakage analysis.} The similar F1-Scores across the two settings suggest that the reported results are not driven by leakage from YouTube.}
\label{tab:youtube_leakage}
\end{center}

\noindent\textbf{Analysis of audio modality in verification.} To investigate the impact of audio information on sentence-wise event verification, we conduct an ANOVA~\cite{st1989analysis} across three audio categories defined by the type of content they convey: Speech (meaningful spoken content), Sound Effect (semantically relevant non-speech sounds), and Background (uninformative audio). As shown in Table~\ref{tab:audio_category}, for Gemini 3 Flash, audio input significantly improves F1-Score in both Speech and Sound Effect, while causing no significant change in the Background. In contrast, F1-Scores of video-SALMONN 2+ decline across these categories. These results suggest that while Gemini effectively leverages audio as a complementary signal, video-SALMONN 2+ tends to treat audio as a distractor.

\begin{table*}[t]
\centering
\resizebox{0.75\textwidth}{!}{%
\begin{tabular}{llccc}
\toprule
Video-LLM & Audio Category & F1-Score Gain (\%) & $t$ & Significant \\
\midrule
\multirow{3}{*}{Gemini 3 Flash} & Speech & +3.3 & +3.9 & *** \\
 & Sound Effect & +4.2 & +3.6 & *** \\
 & Background & $-$0.8 & $-$0.9 & n.s. \\
\midrule
\multirow{3}{*}{video-SALMONN 2+ 7B} & Speech & $-$0.3 & $-$1.7 & n.s. \\
 & Sound Effect & $-$1.5 & $-$2.5 & * \\
 & Background & $-$0.9 & $-$2.7 & ** \\
\midrule
\multirow{3}{*}{video-SALMONN 2+ 72B} & Speech & $-$2.2 & $-$4.2 & *** \\
 & Sound Effect & $-$2.4 & $-$2.6 & * \\
 & Background & $-$1.6 & $-$2.6 & ** \\
\bottomrule
\end{tabular}%
}
\caption{\textbf{F1-Score gain when audio is added by audio category and model.} For each model and audio category, we compute the F1-Score gain (with audio 
- without audio) and test whether this gain is significantly different from zero using a one-sample $t$-test. *$p<.05$, **$p<.01$, ***$p<.001$, n.s.: not significant $p\geq.05$.}
\label{tab:audio_category}
\end{table*}

\end{document}